\documentclass{article}

\PassOptionsToPackage{numbers,sort&compress}{natbib}
\usepackage[preprint]{neurips_2025}

\usepackage[utf8]{inputenc} % allow utf-8 input
\usepackage[T1]{fontenc}    % use 8-bit T1 fonts
\usepackage{hyperref}       % hyperlinks
\usepackage{url}            % simple URL typesetting
\usepackage{booktabs}       % professional-quality tables
\usepackage{amsfonts}       % blackboard math symbols
\usepackage{amsmath}         % math environments
\usepackage{bm}             % bold math symbols
\usepackage{nicefrac}       % compact symbols for 1/2, etc.
\usepackage{microtype}      % microtypography
\usepackage{xcolor}         % colors
\usepackage{graphicx}       % figures
\usepackage{rotating}       % sideways figures
\usepackage{makecell}
\usepackage{multirow}

\newcommand{\cmark}{\checkmark}
\newcommand{\vjepatwo}{\mbox{V-JEPA~2}}
\title{From Pixels to Newtons: Predicting In Vivo Joint Contact Forces from Monocular Video}

\author{%
  Jessy Lauer \\
  \texttt{jlauer@rowland.harvard.edu} \\
}

\begin{document}

\maketitle

\begin{abstract}
  Joint contact forces govern implant longevity, cartilage health, and rehabilitation outcomes---shaping who develops osteoarthritis, who recovers well from joint replacement, and who benefits from biomechanical interventions. Yet they remain measurable only invasively, in a few dozen
  patients worldwide with instrumented implants.
  I present a physics-free pipeline to predict instantaneous 3D hip and knee contact forces from an uncalibrated monocular video---no markers, force plates, electromyography, subject-specific imaging, or musculoskeletal model.
  Parametric body meshes are recovered per frame, encoded as kinematic features, and decoded into forces by a transformer whose pose stream is adaptively modulated at every layer by body shape, joint, side, activity text, and self-supervised video tokens (\vjepatwo{}), unifying hip and knee in a single model. Under leave-one-subject-out cross-validation across 26 patients and 25 activity categories from the \emph{in vivo} OrthoLoad database, the pipeline matches the accuracy of subject-specific musculoskeletal simulations ($0.32 \pm 0.08$\,BW RMSE for hip; $0.23 \pm 0.03$\,BW for knee) and resolves peak force changes smaller than those reported for gait retraining and osteoarthritis progression. Applied zero-shot to an independent instrumented cohort, it rivals or outperforms prior published methods, evidence that the learned mapping transfers beyond its training distribution.
  Even without curated activity labels, video features alone preserve accuracy and enable end-to-end inference on raw footage.
  Driven by the predictor, a generative motion prior produces biomechanically plausible variants with reduced peak loading, independently rediscovering strategies identified in the predictive simulation literature. This pipeline establishes uncalibrated monocular video as a viable modality for estimating joint loading, opening a path toward retrospective analysis of archived clinical recordings, primary-care screening, and at-home rehabilitation tracking.
\end{abstract}

\section{Introduction}

Joint contact forces are the dominant mechanical stimulus to which bone,
cartilage, and the surrounding soft tissues continuously
adapt~\citep{pizzolato2017bioinspired}. The
loading regime of everyday movement (e.g., walking, climbing
stairs, rising from a chair) thus governs both healthy tissue maintenance
and the progression of joint pathology. The assessment of joint
forces is consequently central to orthopedic medicine, from the design and sizing of joint
replacements~\citep{heller2005determination} to the personalized
management of osteoarthritis~\citep{diamond2022feasibility}
and rehabilitation from injury, surgery, or disuse (e.g.,~\citep{gardinier2013altered,diamond2024osteoarthritis}). Despite this recognized clinical importance, joint contact forces
cannot be measured noninvasively. The only direct measurements come from instrumented
implants: prostheses fitted with strain gauges and telemetry that transmit force data in
real time~\citep{dlima2005implantable,bergmann1988multichannel,damm2010total,heinlein2007design}.
Such implants exist in only a few dozen patients worldwide, exclusively post-arthroplasty, leaving
the loading regime of the healthy or pre-surgical joint inaccessible to direct measurement
(e.g., knee:~\citep{dlima2006tibial,fregly2012grand,heinlein2009complete};
hip:~\citep{bergmann2016standardized,rydell1966forces,english1979vivo};
shoulder:~\citep{bergmann2007vivo}). As a
result, the vast majority of clinical and research decisions about joint loading rely on indirect
estimates rather than measurements. Noninvasive estimation of joint loads remains an open challenge in biomechanics, unresolved despite sustained methodological effort~\citep{tomasi2023estimation}.

The dominant indirect approach is \emph{in silico} musculoskeletal
modeling, which represents the individual as a physics-based assembly of
bone segments, idealized joint constraints, and Hill-type muscle-tendon
actuators typically parameterized from cadaver measurements (e.g.,~\citep{rajagopal2016full}).
Given experimental kinematics from optical motion capture and external
kinetics from force plates, muscle and joint contact forces are
estimated by inverse dynamics combined with static optimization or another solution algorithm (e.g.,~\citep{anderson2001static}). Generic models are linearly scaled to a subject's dimensions for basic
personalization~\citep{delp1990interactive}, but scaled models are often poor
representations of an individual's anatomy and yield erroneous contact forces (e.g.,~\citep{wesseling2016subject}); higher-fidelity
personalization---such as MRI-derived bone and joint geometry~\citep{stansfield2026we} or EMG-informed neural control to capture individual
motor strategies static optimization cannot recover (e.g.,~\citep{pizzolato2015ceinms,sartori2014hybrid,lloyd2003emg})---demands substantially more
expertise and instrumentation. Yet modeling choices accumulate at every stage of the pipeline, and small
changes at any one of them can propagate into large differences in
predicted contact forces~\citep{hosseini2022uncertainty,moissenet2017alterations}. Combined with the laboratory
requirements the pipeline inherits (e.g., calibrated multi-camera systems, skin
markers, embedded force plates, and expert operators), these constraints
leave continuous monitoring of joint loading during daily life, where
clinically relevant loading actually accumulates, out of reach.

Monocular human pose estimation has advanced rapidly in recent years. Parametric mesh
recovery from a single RGB camera now approaches multi-view laboratory
accuracy on in-the-wild video (e.g.,~\citep{sarandi2024neural,wang2025prompthmr,yang2026sam3dbody}),
yielding per-frame axis-angle joint rotations and body shape, from which virtual markers and joint locations can be derived. Yet a complete pipeline
from a single uncalibrated camera to \emph{in vivo} joint contact forces
does not exist. Three gaps remain. First, no method is free of
both musculoskeletal simulation and laboratory
instrumentation: video-based approaches that recover kinematics and
ground reaction forces from a smartphone alone (a notable advance removing the need for force plates) still feed a downstream
biomechanical
model~\citep{miller2025integrating,gilon2026opencap,uhlrich2023opencap},
inheriting its modeling assumptions and computational cost;
learned methods that bypass simulation, by contrast, inherit laboratory inputs---marker-based motion
capture, body-worn sensors, force plates, or
EMG~\citep{stetter2019estimation,cornish2024hip,zou2024prediction,chen2026ai}---and typically train at a single joint on a narrow set of activities.
Second, most existing pipelines validate against \emph{in silico} rather
than \emph{in vivo} forces; in the data-driven case the models
are trained on those same simulated targets, so reported accuracies are
bounded above by the simulation they imitate. Third, no prior work has
used a learned force predictor to close the loop on motion design: searching for movement variants that achieve a desired loading
profile.

The OrthoLoad public database~\citep{bergmann2008orthoload} makes it
possible to eliminate all three gaps at once. It pairs synchronized video with
time-resolved 3D bone-to-bone contact forces measured \emph{in vivo} at
the prosthesis across dozens of patients, multiple joints, and
a wide activity repertoire: level walking, stair negotiation, and
sit-to-stand transitions, but also cycling, aquagym, deep knee bends,
and aerobics. I exploit this pairing
to build the first such pipeline, trained and validated entirely on implant measurements rather than simulated targets. Specifically, I make five contributions:

\begin{enumerate}
  \item A physics-free pipeline from uncalibrated monocular video to 3D joint contact forces---no markers, force plates, electromyography, or musculoskeletal simulation---validated under leave-one-subject-out cross-validation against \emph{in vivo} implant recordings from 26 patients across 25 activity categories, with accuracy matching that of laboratory musculoskeletal pipelines.
  \item A single transformer that predicts hip and knee contact forces and per-frame uncertainty from partial per-subject supervision, by adaptively modulating its pose stream at every layer with joint, side, activity, and video context.
  \item Evidence that a frozen video world model (\vjepatwo{}~\citep{assran2025vjepa}), pretrained without biomechanical or semantic supervision, substitutes for curated activity labels at no loss in accuracy, removing a manual labeling bottleneck for clinical deployment.
  \item A closed-loop inverse design procedure that produces biomechanically plausible motion variants with reduced peak loading, steered by the predictor's gradients through a flow matching generative motion prior, and independently rediscovers load reduction strategies identified in the predictive simulation literature.
  \item The first cross-cohort, zero-shot evaluation of joint contact force estimation against out-of-distribution \emph{in vivo} implant data, rivaling past winners of the Grand Challenge competitions~\citep{fregly2012grand}.
\end{enumerate}

Trained model weights and code\footnote{\url{https://github.com/jeylau/jcf}} will be released publicly, along with scripts to reproduce the processed dataset (SMPL pose sequences, motion features, aligned force signals, and activity labels) from OrthoLoad, as a benchmark for video-based biomechanical analysis. A companion web interface will provide cloud-based inference on user-uploaded video, lowering the deployment barrier for non-technical users.

\section{Methods}

\begin{figure}[!htbp]
  \centering
  \includegraphics[width=\linewidth]{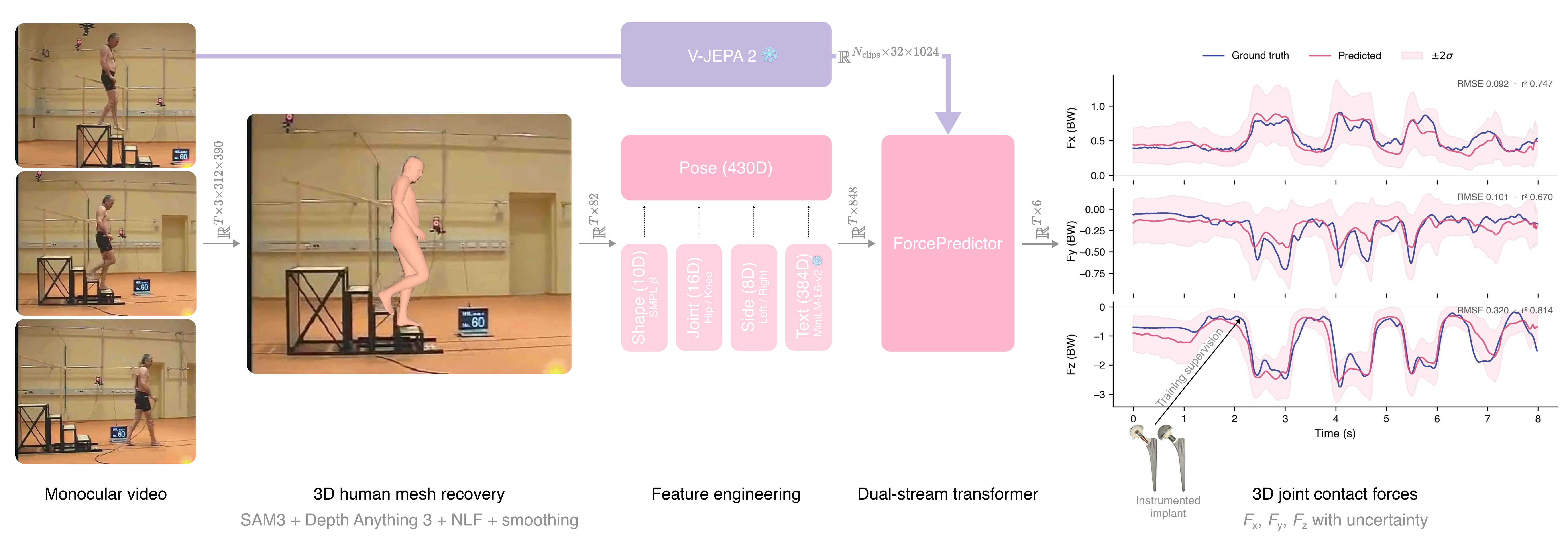}
  \caption{Overview of the proposed methodology for predicting 3D instantaneous joint contact forces from uncalibrated monocular video. Each clip is processed by SAM\,3 for person detection and tracking, Depth Anything\,3 for camera pose estimation, followed by NLF-based SMPL mesh recovery and an iterative smoothing procedure; the recovered pose is converted into a 430-dimensional motion feature vector and combined with static conditioning tokens encoding SMPL shape $\bm{\beta}$, target joint, implant side, and a frozen MiniLM-L6-v2 sentence embedding of the activity label. A dual-stream transformer fuses this pose stream with auxiliary tokens from a frozen \vjepatwo{} video encoder via gated cross-attention, and heteroscedastic output heads produce per-frame mean and uncertainty estimates of the three force components $(F_x, F_y, F_z)$ in the bone-based coordinate system, normalized by body weight. Ground-truth forces, used only for supervision, are recorded by the instrumented implant via telemetry; the rightmost stage shows predicted means (rose), $\pm 2\sigma$ uncertainty bands (shaded), and \emph{in vivo} implant measurements (indigo) for a representative held-out sequence. $T$ denotes sequence length (set to 256 during training; the full video length at inference).}
  \label{fig:workflow}
\end{figure}

The complete pipeline (from raw monocular video to instantaneous force prediction) is summarized in
Fig.~\ref{fig:workflow}; each stage is detailed in the subsections below.

\subsection{Dataset}

\begin{figure}[!htbp]
  \centering
  \includegraphics[width=\linewidth]{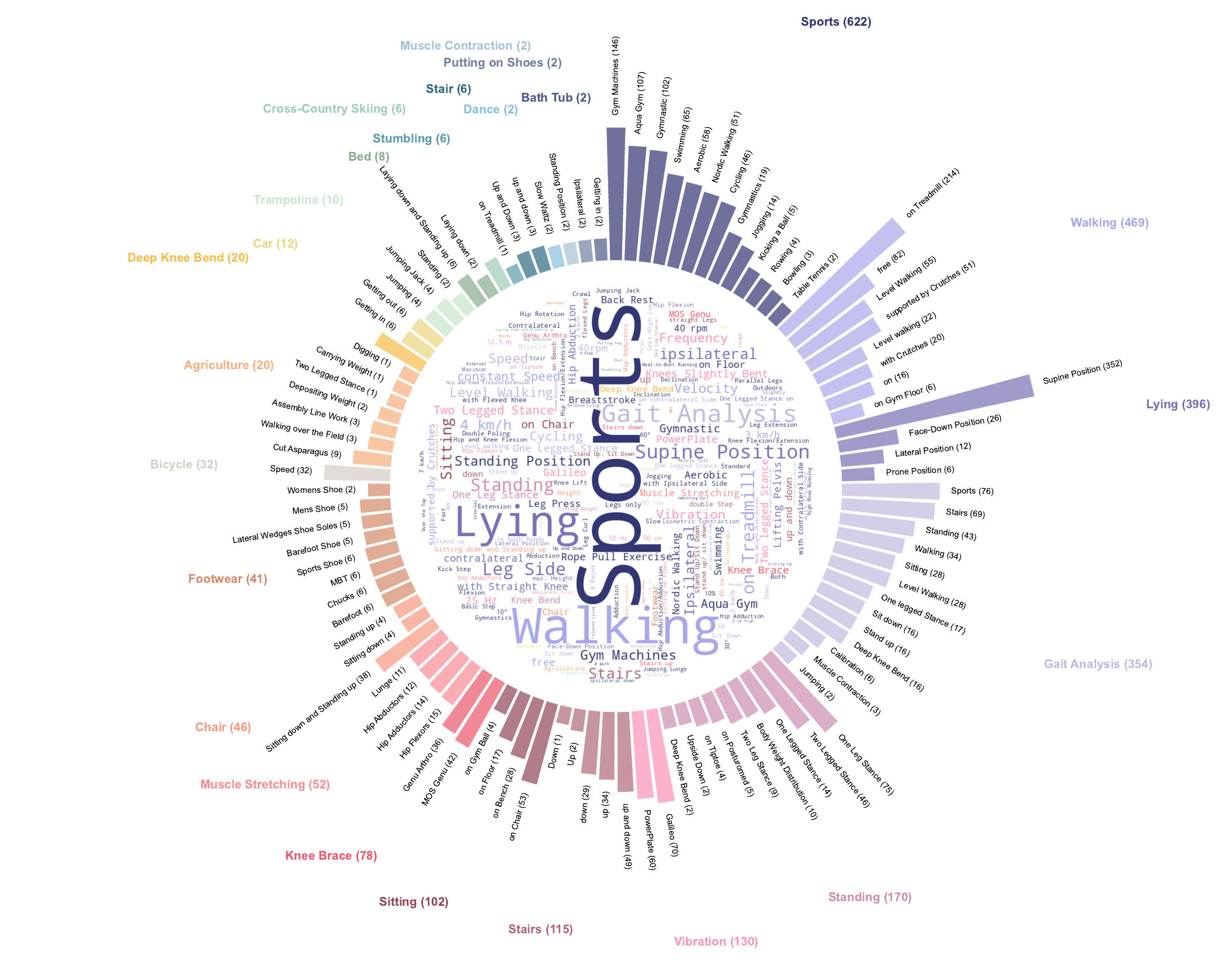}
  \caption{Distribution of activity labels in the OrthoLoad dataset. Radial bars represent individual sub-activities, with height proportional to sample count (square-root scaled) and color denoting the
    parent category. Top-level categories are labeled at the outer periphery with their total counts. The central word cloud displays all activity terms sized by frequency across the dataset,
    colored consistently with their parent category. The taxonomy comprises 25 top-level categories and their constituent sub-activities, ranging from gait analysis and sports to flexibility exercises, daily-living tasks, and resting postures, reflecting the diversity of \emph{in vivo} loading scenarios captured by instrumented implants.}
  \label{fig:activities}
\end{figure}

All data were obtained from the OrthoLoad database~\citep{bergmann2008orthoload}, a public repository of
synchronized video and \emph{in vivo} joint contact force recordings from patients with instrumented
hip and knee implants. Each recording pairs a short video of a functional activity (mean duration 11.3\,s; range 0.7--66.1\,s) with time-resolved 3D force vectors sampled at the implant at
approximately 200\,Hz. Force components are expressed in a bone-based coordinate system and
normalized by body weight (BW); for knee implants, forces originally reported in an implant-based
coordinate system were rotated into this frame using the implant alignment angles stored in each
recording's header. Activity labels follow a hierarchical string
format (e.g., ``Walking~>~with Crutches~>~on Contralateral Side''); where multiple labels existed
for the same recording, only the most specific was retained. The retained labels resolve into 25 top-level activity categories, each comprising multiple sub-activities; their distribution is shown in Fig.~\ref{fig:activities}.

Each patient is instrumented at a single joint type (hip or knee); two carry bilateral hip implants, one per side. A subset of 146
trials were recorded with two synchronized camera views (left and right) displayed side by side;
these were split into separate video files prior to pose estimation, yielding 2,843 distinct videos (recorded at 25 or 50\,fps).
After excluding shoulder recordings (too few samples; $n = 140$) and removing 103 sequences with
poor pose reconstruction quality, the final dataset comprised 2,600 video--force pairs (2,003 hip; 597 knee) from 28 instrumented implants across 26 patients.
Patient metadata (body mass, joint type and implant side) were extracted alongside each recording: body mass normalizes the ground-truth forces to body weight, while implant side and joint type serve as model conditioning variables. Cohort characteristics are provided in
Table~\ref{tab:patients}.

\begin{table}[t]
  \centering
  \caption{Cohort characteristics. Each row is an instrumented implant (28 implants, 26 patients). Patients EB and KW carry bilateral hip implants and therefore appear as two rows each. Sex, age, mass, height, and indication are properties of the host patient. Side: L\,=\,left, R\,=\,right. Sex: m\,=\,male, f\,=\,female. Age at time of implantation. $n$: number of recorded trials. Dashes indicate unavailable data.}
  \label{tab:patients}
  \small
  \setlength{\tabcolsep}{6pt}
  \begin{tabular}{llccrrrrl}
    \toprule
    Implant & Joint & Side & Sex & Age (yr) & Mass (kg) & Height (m) & $n$ & Indication            \\
    \midrule
    EBL     & Hip   & L    & m   & 83       & 62        & 1.68       & 279 & Osteoarthritis        \\
    EBR     & Hip   & R    & m   & 83       & 62        & 1.68       & 31  & Osteoarthritis        \\
    IBL     & Hip   & L    & f   & 76       & 84        & 1.70       & 65  & Osteoarthritis        \\
    JBR     & Hip   & R    & f   & 69       & 47        & 1.60       & 45  & Femoral head necrosis \\
    HSR     & Hip   & R    & m   & 55       & 82        & 1.74       & 106 & Osteoarthritis        \\
    KWR     & Hip   & R    & m   & 61       & 72        & 1.65       & 128 & Osteoarthritis        \\
    KWL     & Hip   & L    & m   & 61       & 72        & 1.65       & 22  & Osteoarthritis        \\
    PFL     & Hip   & L    & m   & 49       & 98        & 1.75       & 103 & Osteoarthritis        \\
    RHR     & Hip   & R    & f   & 63       & 60        & ---        & 26  & Osteoarthritis        \\
    H1L     & Hip   & L    & m   & 55       & 73        & 1.78       & 18  & Coxarthrosis          \\
    H2R     & Hip   & R    & m   & 61       & 75        & 1.72       & 226 & Coxarthrosis          \\
    H3L     & Hip   & L    & m   & 59       & 92        & 1.68       & 107 & Coxarthrosis          \\
    H4L     & Hip   & L    & m   & 50       & 85        & 1.78       & 108 & Coxarthrosis          \\
    H5L     & Hip   & L    & f   & 62       & 87        & 1.68       & 114 & Coxarthrosis          \\
    H6R     & Hip   & R    & m   & 68       & 84        & 1.76       & 171 & Coxarthrosis          \\
    H7R     & Hip   & R    & m   & 52       & 95        & 1.79       & 186 & Coxarthrosis          \\
    H8L     & Hip   & L    & m   & 55       & 80        & 1.78       & 106 & Coxarthrosis          \\
    H9L     & Hip   & L    & m   & 54       & 118       & 1.81       & 103 & Coxarthrosis          \\
    H10R    & Hip   & R    & f   & 53       & 98        & 1.62       & 59  & Coxarthrosis          \\
    \midrule
    K1L     & Knee  & L    & m   & 63       & 100       & 1.77       & 94  & Osteoarthritis        \\
    K2L     & Knee  & L    & m   & 71       & 93        & 1.71       & 64  & Osteoarthritis        \\
    K3R     & Knee  & R    & m   & 70       & 95        & 1.75       & 88  & Osteoarthritis        \\
    K4R     & Knee  & R    & f   & 63       & 92        & 1.70       & 24  & Osteoarthritis        \\
    K5R     & Knee  & R    & m   & 60       & 94        & 1.75       & 136 & Osteoarthritis        \\
    K6L     & Knee  & L    & f   & 65       & 76        & 1.74       & 18  & Osteoarthritis        \\
    K7L     & Knee  & L    & f   & 74       & 70        & 1.66       & 47  & Osteoarthritis        \\
    K8L     & Knee  & L    & m   & 70       & 77        & 1.74       & 86  & Osteoarthritis        \\
    K9L     & Knee  & L    & m   & 75       & 100       & 1.66       & 40  & Osteoarthritis        \\
    \bottomrule
  \end{tabular}
\end{table}

\subsection{Pose estimation and temporal smoothing}

Full-body 3D pose was recovered from each monocular video using Neural Localizer Fields (NLF~\citep{sarandi2024neural}), a fast state-of-the-art method
that estimates SMPL~\citep{loper2023smpl} body mesh parameters (pose, shape, and global translation) for
every frame. Because roughly half of the OrthoLoad videos contain multiple people (therapists,
experimenters) and occasionally subjects in lying positions missed by standard person detectors,
SAM\,3~\citep{carion2025sam} with text prompts was used for detection and tracking prior to mesh recovery;
bounding boxes were scaled by a factor of 1.2 to provide sufficient context for the mesh estimator.

\paragraph{Camera parameter estimation.}
The OrthoLoad videos are recorded with uncalibrated cameras, and some recordings involve camera
motion. To provide the SMPL fitting stage with accurate perspective geometry, per-frame camera
extrinsics and intrinsics were estimated from the video using Depth Anything\,3~\citep{lin2025depth} (DA3-Giant). Uniformly spaced keyframes were passed to DA3, which jointly predicts
metric monocular depth, camera extrinsic matrices
$[\mathbf{R} \mid \mathbf{t}] \in \mathbb{R}^{3 \times 4}$, and intrinsic matrices
$\mathbf{K} \in \mathbb{R}^{3 \times 3}$. Extrinsics were interpolated to all video frames using
spherical linear interpolation for rotations and piecewise cubic Hermite interpolation for translations, followed by robust temporal smoothing via iteratively reweighted least
squares with a trapezoidal kernel ($\sigma = 0.4$\,s). Intrinsics were interpolated similarly, with focal lengths in
log space to preserve scale consistency. The resulting
per-frame camera parameters were passed to NLF, enabling perspective-correct SMPL pose estimation.

Despite NLF's strong performance on in-the-wild benchmarks, the raw per-frame SMPL fits exhibited
temporal jitter and occasional large outliers, common artifacts of monocular
estimation in the absence of temporal cues. A multi-stage robust temporal smoothing pipeline was therefore applied.\footnote{This procedure
  follows the post-processing code released by the NLF authors at \url{https://github.com/isarandi/nlf-pipeline}.} First, the SMPL
body model was refit to the detected mesh vertices with a shared shape vector $\bm{\beta}$
across all frames of a sequence, enforcing anthropometric consistency; a median-based scale
correction was applied simultaneously. Next, the root joint trajectory was smoothed with a
large-kernel iteratively reweighted least squares (IRLS) filter using Gaussian weights, which also
served to detect shot-cut discontinuities in videos with camera transitions. A second, smaller-kernel
IRLS pass then smoothed all joint trajectories. Last, the SMPL model was refit a final time with a
shared $\bm{\beta}$ to the smoothed vertices, and root translation was smoothed once more to
remove any residual high-frequency drift.

\subsection{Motion feature representation}

To ensure a consistent coordinate frame, the first-frame yaw rotation was removed so that all
sequences begin with the subject facing a canonical direction; for sequences in which the subject was
lying down, the head-to-toe axis was used instead. The horizontal origin was set to the first-frame
root position. Crucially, yaw was only removed from the first frame: subsequent frames retain the
original heading changes, preserving rotational dynamics (e.g., turning during gait) that may affect
joint loading.

Following the motion representation adopted in STMC~\citep{petrovich2024stmc}, each frame is represented by a
430-dimensional vector $\bm{x} \in \mathbb{R}^{430}$, extended here with joint linear and
angular velocity terms that proved critical for force prediction:
\[
  \bm{x} \;=\; [\dot{\bm{\alpha}}_y,\; \dot{\bm{r}}_{xz},\; r_y,\;
  \bm{\theta},\; \bm{j},\; \dot{\bm{j}},\; \dot{\bm{\theta}}],
\]
where $\dot{\bm{\alpha}}_y \in \mathbb{R}^{1}$ is the root angular velocity about the vertical
axis, $\dot{\bm{r}}_{xz} \in \mathbb{R}^{2}$ the root linear velocity in the body-local horizontal
plane, and $r_y \in \mathbb{R}^{1}$ the root height. $\bm{\theta} \in \mathbb{R}^{144}$ denotes the
SMPL pose parameters for 24 joints encoded using the continuous 6D rotation
representation~\citep{zhou2019continuity}, and $\bm{j} \in \mathbb{R}^{69}$ the 3D positions of the
23 non-root joints expressed relative to the root in a body-local frame, yielding a
rotation-invariant spatial description. $\dot{\bm{j}} \in \mathbb{R}^{69}$ and
$\dot{\bm{\theta}} \in \mathbb{R}^{144}$ are the first-order finite differences of joint positions
and 6D rotations, divided by the inter-frame interval so that velocities are expressed per unit time and remain comparable across the 25 and 50\,fps recordings; feature ablation experiments confirmed that these velocity terms are
among the most informative inputs for force prediction.

All features were z-score normalized using statistics computed from valid frames of the training set.
To prevent near-constant features from dominating after normalization (a risk when individual
standard deviations are very small), features within each semantic group (e.g., all 6D rotation
dimensions, all relative-position dimensions) were assigned the group-mean standard deviation rather
than their individual values. This preserved relative magnitudes within a group while stabilizing the
scale across groups.

Second-order temporal features (linear and angular accelerations $\ddot{\bm{j}}$, $\ddot{\bm{\theta}}$) were evaluated but excluded. While Newton's
second law motivates their inclusion, the second-order differentiation needed to estimate them amplified pose estimation noise, and ablation experiments showed no improvement in validation nRMSE.

\subsection{Force preprocessing}

Ground-truth force signals, sampled at approximately 200\,Hz, were resampled to the video frame rate via linear interpolation and normalized by
each patient's body weight to yield dimensionless forces in units of body weight (BW). For knee recordings, forces were rotated from the
implant-based coordinate system to the bone-based system using the implant alignment angles provided in each file's header. Hip I/II and Hip
III implants report forces in left-femur and right-femur coordinate systems, respectively. While the $x$-axis convention differs (medial versus lateral), the documented sign convention for Hip I/II ($-F_x, -F_y, -F_z$) compensates for this, so that $F_x$ requires no correction.
$F_y$ and $F_z$ were negated for Hip I/II recordings to align them with the Hip III convention, ensuring that all force components---medial--lateral ($F_x$),
anterior--posterior ($F_y$), and proximal--distal ($F_z$)---share a consistent anatomical interpretation across implant generations.

\subsection{Force prediction model}

The force prediction model is a transformer that maps motion features to per-frame 3D force predictions.
Its pose stream is conditioned at every layer by joint, side, morphology, and
activity labels, and, when video is available, cross-attends to self-supervised
\vjepatwo{} tokens. A single model handles both hip and knee, enabling cross-joint transfer from the
full dataset rather than training separate, data-starved models per joint type. Per-frame mean and
log-variance heads yield heteroscedastic predictions of the three force components.

\paragraph{Input projection.}
Each training sample is a random $T = 256$-frame crop of a motion sequence, a temporal
augmentation that exposes the model to diverse segments across epochs.
At each frame, the 430-dimensional motion feature vector is concatenated with a 10-dimensional SMPL
shape vector $\bm{\beta}$, a 384-dimensional text embedding of the activity label obtained from
a frozen all-MiniLM-L6-v2 sentence encoder and z-scored per dimension, a
16-dimensional learned joint-type embedding, and an 8-dimensional learned implant-side embedding
(left or right). The concatenated vector is projected to the model dimension $d = 256$ by a single
linear layer. Using a pretrained sentence encoder provides semantic similarity between related
activities (e.g., ``Walking'' and ``Walking with Crutches'') without requiring enough samples per
category for learned embeddings to converge.

\paragraph{Local temporal convolutions.}
Before entering the transformer, tokens pass through two 1D convolutional layers (kernel size~5) with
a residual connection. These local filters capture short-range temporal dynamics (e.g., onset slopes,
jerk) that are central to force generation and that full self-attention would need multiple layers to
model.

\paragraph{Rotary positional encoding.}
Rotary Position Embeddings (RoPE~\citep{heo2024rotary}) are used to encode temporal position. Rather than adding a fixed positional signal to the input, RoPE applies position-dependent rotations to the query and
key vectors within each attention head, encoding relative temporal distance directly into the attention
logits. For self-attention over pose tokens, positions correspond to frame indices; for cross-attention to
\vjepatwo{} video tokens, the keys are rotated according to their source frame positions, ensuring temporal
alignment between modalities.

\paragraph{Adaptive layer normalization for static conditioning.}
Subject morphology, activity type, target joint, and implant side are static within each sample but critically influence how kinematics map to joint reaction forces. To provide a multiplicative conditioning pathway at every transformer layer, Adaptive Layer Normalization (AdaLN)~\citep{peebles2023scalable} is used. A small MLP encodes the concatenated static features (SMPL shape parameters, sentence embedding of the activity label, and learned joint and side embeddings) into a conditioning vector, which produces per-layer scale and shift parameters that modulate the layer-normalized activations before each sublayer. The static features are also concatenated with the per-frame pose features at the input, so that the convolutional stem retains direct access to the full context. The modulation parameters are zero-initialized so that the network starts as a standard transformer and
gradually learns to incorporate conditioning.

\paragraph{Temporal backbone.}
The token sequence is processed by a transformer encoder.
In the base configuration, this is a standard multi-head self-attention encoder. An extended variant
replaces the encoder with a stack of gated decoder layers that perform self-attention over the pose stream
and cross-attention to temporally aligned video tokens from \vjepatwo{}~\citep{assran2025vjepa}, a self-supervised
video encoder pretrained on over a million hours of internet video,
with strong motion understanding and state-of-the-art action
anticipation performance. Each video was divided into non-overlapping
64-frame clips, yielding up to $\lceil T/64 \rceil + 1 = 5$ clips per training sample (4 when the crop aligns to a clip boundary); the encoder produced 32 temporal tokens per clip (tubelet size~2, dimension~1024),
spatially pooled via global average over $16 \times 16 = 256$ patches. At training time, tokens were aligned to pose frames
using their source frame positions via RoPE. Each
cross-attention module includes a learnable scalar gate initialized to zero, so that the model begins
training as a pose-only network and gradually incorporates video context; this prevents random
cross-attention projections from corrupting the learned pose representations early in training.
The force predictor uses 6 transformer layers with 8 attention heads and model dimension $d = 256$ (6.2M parameters). Dropout of 0.1 was applied throughout.

\paragraph{Output heads.}
Two parallel output heads produce, for each frame, a mean force vector
$\bm{\mu} \in \mathbb{R}^3$ and a log-variance vector
$\log \bm{\sigma}^2 \in \mathbb{R}^3$. The mean head is a single linear projection; the
log-variance head is a two-layer MLP with GELU activation, giving it additional capacity to model
input-dependent uncertainty. Together these define a heteroscedastic Gaussian over the three force
components, allowing the model to express per-frame, per-axis uncertainty. The log-variance head's
output bias is initialized to $-2.0$, corresponding to a prior standard deviation of approximately
0.37\,BW.

\subsection{Training procedure}

\paragraph{Loss functions and staged training.}
Training proceeds in three stages. In Stage~1, the full model---including, when used, the gated cross-attention layers over \vjepatwo{} features (gates initialized to zero)---is trained end-to-end with
a masked mean squared error loss on valid frames. In Stage~2,
all parameters except the log-variance head are frozen, and the model is fine-tuned with the $\beta$-NLL loss~\citep{seitzer2022pitfalls}, which weights each sample by the detached $\sigma^{2\beta}$ ($\beta = 0.5$) to prevent the variance from collapsing or exploding. In Stage~3, all parameters are unfrozen and training continues with $\beta$-NLL loss at a reduced learning rate.

\paragraph{Data split and cross-validation.}
All reported generalization metrics use leave-one-subject-out cross-validation (LOSO CV) over the 26 patients. Each fold holds out one patient (all of their trials, and both implants for the bilaterally instrumented patients EB and KW) and selects the best epoch on a validation split drawn from the remaining 25. This protocol ensures that reported metrics reflect truly out-of-sample performance, with no information leakage from the test patient into training or model selection. Ablation studies, for which a full LOSO sweep is prohibitive, instead use a single 85/15 patient-level split stratified by joint type to preserve the hip/knee ratio, with no patient appearing in both sets. For inverse design, a single model is trained on all available patients using the same architecture and hyperparameters. Because the optimization targets motion trajectories of patients already present in the dataset, training on the full cohort yields the strongest differentiable surrogate without compromising the generalization analysis established by LOSO CV.

\paragraph{Optimizer.}
The force predictor was trained with AdamW~\citep{loshchilov2017adamw} (learning rate $10^{-4}$, weight decay
$10^{-2}$, batch size 256) with cosine annealing and gradient clipping (max norm 1.0).
Stage~1 ran for 100 epochs, Stage~2 (variance head only) for 20 epochs, and Stage~3
(full model, reduced learning rate $10^{-6}$) for 50 epochs.

\subsection{Post-hoc uncertainty calibration}
\label{sec:methods_calibration}

Although Stage~2 training shapes $\hat{\bm{\sigma}}$ to track
input-dependent heteroscedasticity, the $\beta$-NLL objective does not
guarantee that the absolute scale of $\hat{\sigma}$ is calibrated to
residual magnitudes. Per-axis multiplicative
temperature scaling~\citep{levi2022evaluating} is therefore applied as a post-hoc step:
\[
  \hat{\sigma}^{\mathrm{cal}}_{t,a} \;=\; \tau_a \cdot \hat{\sigma}_{t,a},
  \qquad a \in \{x, y, z\}.
\]
The scalars $\tau_a$ are fitted by maximum likelihood under a Gaussian
likelihood, which admits the closed form
\[
  \tau_a^2 \;=\; \frac{1}{N} \sum_t
  \Big( (y_{t,a} - \hat{\mu}_{t,a}) / \hat{\sigma}_{t,a} \Big)^2.
\]
To prevent leakage, $\tau_a$ is fitted in a leave-one-fold-out manner:
for each held-out patient, the calibration constant is computed from
residuals of the remaining LOSO folds only.
Because $\tau_a$ is multiplicative and shared across all frames of a
given axis, this step preserves the relative within-trial structure of
$\hat{\sigma}$ and modifies only the absolute scale of the predictive
bands. All reported coverage statistics and uncertainty-aware
quantities (Sec.~\ref{sec:inverse_design}) use the calibrated
$\hat{\sigma}^{\mathrm{cal}}$.

\subsection{Inverse design}
\label{sec:inverse_design}

Beyond prediction, the goal was to identify motion variants that reduce peak joint loading while remaining biomechanically plausible: an inverse design objective. The approach combines a generative motion prior with gradient-based guidance from the trained force predictor.

\paragraph{Motion prior: conditional flow matching.}
A generative model was trained on the space of SMPL 6D joint rotation sequences using rectified
flow~\citep{lipman2022flow,liu2022flow}, a conditional flow matching framework in which a learned
velocity field $v_\theta(\mathbf{x}_t, t)$ transports samples from a standard Gaussian prior to the
data distribution along straight-line paths. Specifically, training pairs were constructed as
$\mathbf{x}_t = (1 - t)\,\mathbf{x}_0 + t\,\bm{\epsilon}$, $\bm{\epsilon} \sim
  \mathcal{N}(\mathbf{0}, \mathbf{I})$, and the network was trained to predict the velocity
$\mathbf{v} = \bm{\epsilon} - \mathbf{x}_0$. The velocity field was parameterized by a
Diffusion Transformer (DiT)~\citep{peebles2023scalable} with adaptive layer normalization (adaLN-Zero): each
transformer block receives per-layer scale, shift, and gate parameters conditioned on the diffusion
timestep, allowing fine-grained temporal modulation of the denoising dynamics.
The flow model comprises 4 adaLN-Zero transformer blocks with 8 attention heads and dimension $d = 256$ (5.1M parameters), operating on windows of 64 frames of 144-dimensional 6D rotations. It was trained for
1{,}000 epochs with AdamW (learning rate $3 \times 10^{-4}$, no dropout) and gradient clipping (max
norm 1.0) on the same training split as the force predictor, using the standard rectified flow MSE
objective. An exponential moving average (EMA) of model weights was maintained with a target decay of
0.999, ramped up during early training via $\tilde{\gamma} = \min(\gamma,\, (1 + e) / (10 + e))$
where $e$ is the epoch index; the EMA copy was used for all generation.

\paragraph{Guided generation via SDEdit.}
To produce motion variants close to a reference sequence, an SDEdit strategy~\citep{meng2021sdedit} was adopted: the original motion was partially noised to a chosen start time $t_\text{start} < 1$ and then denoised via ODE
integration of the learned velocity field back to $t = 0$. At each integration step, the current state
$\mathbf{x}_t$ was projected onto the clean data manifold via $\hat{\mathbf{x}}_0 = \mathbf{x}_t - t\,\mathbf{v}_\theta$, and the force predictor's gradient with respect to this denoised estimate was computed
for a chosen force objective and added to the velocity field, steering the trajectory toward lower-loading
solutions. Gradients were normalized before application to prevent instability. A cosine ODE schedule
concentrated integration steps near $t = 0$, where small perturbations have the largest effect on
output fidelity. The multiplication by the integration step size $\mathrm{d}t$, which decreases near
$t = 0$ under the cosine schedule, implicitly scales guidance strength by noise level, analogous to
the noise-dependent weighting in Diffusion Posterior Sampling~\citep{chung2022diffusion}.

\paragraph{Force objectives.}
Several differentiable objectives were implemented, selectable per design query: peak absolute force
along any single axis ($F_x$, $F_y$, or $F_z$), computed via a log-sum-exp soft maximum for
smoothness; mean compressive load (time-averaged $F_z$); impulse (time integral of $|F_z|$); and
peak resultant force magnitude. Each objective can target either the hip or knee joint via the
model's joint-conditioning mechanism, regardless of which joint provided the original training
signal. To exploit the calibrated uncertainty estimates (Sec.~\ref{sec:methods_calibration}), an uncertainty-aware variant of the force
objective replaces $|F_z|$ with the upper confidence bound $|F_z| + k\sigma_z$, where $\sigma_z$ is
the predicted per-frame standard deviation and $k$ controls the confidence level. This steers the
optimization toward motions that are predicted to have low loading with high confidence,
avoiding regions of the force predictor's input space where nominal reductions may reflect model
uncertainty rather than genuine biomechanical improvement.

The use of first-order features only was further motivated by the inverse design setting: when acceleration features were included, gradient guidance
exploited small perturbations in second-derivative space to achieve nominal force reductions without meaningful kinematic change: artifacts invisible to the generative prior, which operates on joint rotations rather than positional derivatives.

\paragraph{Optimization setup.}
For each design query, the rectified flow is integrated with $N = 50$ ODE
steps on the cosine schedule from $t_\text{start}$ down to $t = 0$. At
every step the predictor's gradient with respect to the denoised estimate
$\hat{\mathbf{x}}_0$ is rescaled to unit Frobenius norm and added to the
velocity field with a fixed weight $\lambda = 10$; the implicit cosine-schedule
$\mathrm{d}t$ scaling described above then up-weights guidance at low noise
levels. Trials longer than the model window ($T > W = 64$ frames at 25\,Hz)
are cropped to a $W$-frame window centered on the predictor's argmax-$|F_z|$
frame, so that the optimization always operates on the segment containing
peak loading; trials with $T < W$ are excluded
($77/2{,}600 = 3.0\%$ of trials in the full split). Larger $t_\text{start}$
corresponds to a noisier initialization and therefore a larger admissible
edit; $t_\text{start}$ is swept over $\{0.10, 0.15, 0.20, 0.25, 0.30\}$ to
trace out the trade-off between motion change and force reduction. Each
$(t_\text{start}, \text{trial})$ pair is solved under three independent
random initializations $\bm{\epsilon}$ of the noised state; within a seed,
$\bm{\epsilon}$ is shared across $t_\text{start}$ values to give smooth
single-seed trajectories, while across seeds it varies independently to
expose optimizer variance. Reductions are reported as $\max(0,
  (F_z^{\text{orig}} - F_z^{\text{opt}})/F_z^{\text{orig}})$: trials in which
the optimizer increased predicted peak force (i.e.\ failed to find a lower
solution under the chosen $t_\text{start}$ and seed) are reported as zero
reduction rather than negative.

\paragraph{Plausibility check.}
A common failure mode of gradient-guided generation is to push the model
into adversarial regions of input space where the predictor reports
spuriously low loads with collapsed uncertainty. As an orthogonal sanity
check independent of the flow prior, the ratio of post- to
pre-optimization predictive standard deviation,
$\bar{\sigma}_\text{opt}/\bar{\sigma}_\text{orig}$, is reported per axis
as the mean over frames. Values close to one indicate that the optimized
motion lies in a region where the predictor's reported uncertainty is
comparable to that on the original motion; values $\gg 1$ would flag
adversarial regions where nominal force reductions coincide with predictor
confidence collapse. This metric is invariant to per-axis temperature
scaling and therefore independent of the calibration choice in
Sec.~\ref{sec:methods_calibration}. Optimized motions that do not inflate
the predictor's uncertainty are taken as evidence that the predicted
reductions reflect genuine biomechanical strategies rather than gradient
exploits.

\subsection{Evaluation metrics}
Prediction quality was assessed using root mean square error (RMSE) per force component and overall, normalized RMSE (nRMSE, RMSE divided by the peak ground-truth force magnitude of each trial), and the squared Pearson correlation ($r^2$) per component. All metrics were computed on valid frames only, excluding padded regions, and reported per joint type (hip, knee), per implant, and per activity category. Because the per-trial error distributions are right-skewed, results are summarized by the median and interquartile range (IQR)---taken over trials or over implants, as indicated---rather than the mean; the per-implant mean and standard deviation are reported only in Table~\ref{tab:accuracy-comparison}, to match the convention of the prior work compared there.
Predictive uncertainty is evaluated by empirical coverage at
$\pm 2\hat{\sigma}^{\mathrm{cal}}$, computed as the per-axis fraction
of valid frames where $|y_{t,a} - \hat{\mu}_{t,a}| \leq
  2\hat{\sigma}^{\mathrm{cal}}_{t,a}$, averaged across LOSO held-out
trials.

\paragraph{\vjepatwo{} versus text modality comparison.}
Three model variants were compared---a baseline (kinematics + shape), a
text-augmented variant (baseline + activity-label text embeddings), and a
video-augmented variant (baseline + \vjepatwo{} features)---all sharing
identical architecture, training data, and hyperparameters, on the held-out
validation split. For each trial, the per-sample improvement in
nRMSE conferred by each auxiliary modality relative to baseline was computed, $\Delta_{\text{text}} = \mathrm{nRMSE}_{\text{base}} - \mathrm{nRMSE}_{\text{+text}}$ and $\Delta_{\text{video}}$ analogously, and their linear association across all trials was quantified by the Pearson correlation coefficient. Trials were stratified into 14 activity categories; for each category with $n \geq 6$
trials ($n = 11$), the paired difference $\Delta = \mathrm{nRMSE}_{\text{+text}} - \mathrm{nRMSE}_{\text{+video}}$ was characterized by a 95\% percentile bootstrap confidence interval (2,000 resamples with replacement). Categories whose interval excluded zero are reported as favoring the corresponding modality; these intervals are uncorrected for multiple comparisons and are interpreted descriptively, with the direction of the effects across categories as the result of interest.

\paragraph{MDC\textsubscript{95} computation.}
The computation was restricted to gait analysis (walking and stair negotiation trials specifically), the
two activities for which clinical thresholds for changes in peak hip
and knee contact force are well-established in the gait retraining and
arthroplasty literature~(e.g.,~\citep{price2017reliability,cornish2024hip}). For each LOSO held-out trial, per-cycle peak
axial force was extracted from the ground-truth trace using
\texttt{scipy.signal.find\_peaks}.
For each detected ground-truth cycle peak, the
matched predicted peak was taken within a $\pm 0.2$\,s window,
accommodating small phase shifts between predicted and measured force traces.
Per-trial peak prediction error was defined as the signed cycle mean
difference between predicted and measured peaks (in BW).
A linear mixed-effects model was then fit on per-trial errors with a
random subject intercept (\texttt{statsmodels.mixedlm}), separately for
walking and for stair negotiation. The random subject intercept partials
out per-subject prediction biases that cancel when comparing two
measurements of the same subject. The residual scale $\hat{\sigma}$ was taken as the standard deviation
of the LME residuals.
The (two-sided) minimum detectable change at 95\% confidence was then computed as
$\mathrm{MDC}_{95} = \sqrt{2}\,z_{0.975}\,\hat{\sigma} \approx 2.77\,\hat{\sigma}$,
where the $\sqrt{2}$ factor accounts for the variance of the difference
between two independent same-subject measurements~\citep{weir2005quantifying}.
$\mathrm{MDC}_{95}$ is reported in absolute units (BW) and relative to the
trial-mean ground-truth peak (\%).

\subsection{External validation on the Grand Challenge dataset}
\label{sec:methods-grand-challenge}

The model was evaluated on the Grand Challenge Competitions to Predict
In Vivo Knee Loads~\citep{fregly2012grand}, comprising four subjects across six competitions, each with force-measuring tibial prostheses and synchronized Vicon marker trajectories. All
195 trials were included. Predictions were fully blinded: no data from the competitions appeared in training, and the
OrthoLoad-trained model was applied as-is, without fine-tuning or
per-subject recalibration. SMPL body meshes were recovered from the
marker trajectories using MoSh++~\citep{mahmood2019amass}, which
solves for per-frame SMPL pose and shape parameters from the marker
trajectories. Each trial was assigned
a label from the OrthoLoad activity vocabulary used during training
(e.g., ``Gait Analysis;~>~Level Walking;'' or ``Walking;~>~on Treadmill;~>~constant Speed;~>~3 km/h;''),
with treadmill velocities taken from the competition protocol. Implant
forces were converted from pounds to body weights; no basis change was
applied between the tibial-tray frame in which the implant reports
forces and the tibia-based frame used by the OrthoLoad knee training
data, since no robust alignment between the two could be determined.
In the absence of synchronized video, inference used the full model
with the \vjepatwo{} cross-attention pathway zeroed out.

\section{Results}

\subsection{Accurate per-frame force prediction across loading regimes}

\begin{sidewaysfigure}
  \centering
  \includegraphics[width=\textheight]{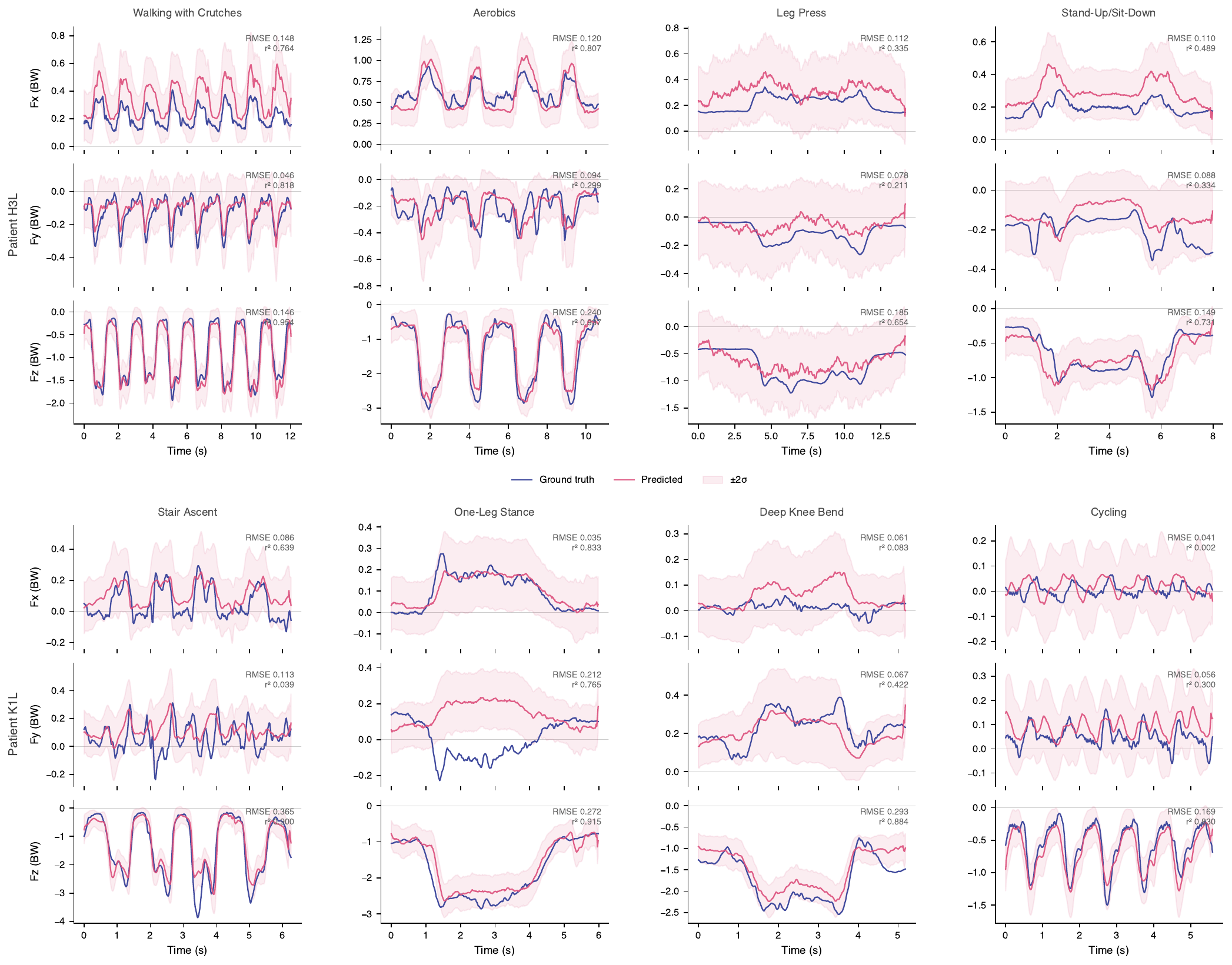}
  \caption{Predicted versus ground-truth joint contact forces for two cases held out during leave-one-subject-out cross-validation (H3L, top; K1L, bottom). Each column shows one trial, with rows corresponding to the three force components ($F_x$, $F_y$, $F_z$) expressed in body weights (BW). Shaded regions denote the $\pm 2\sigma$ predictive uncertainty. Per-component RMSE and $r^2$ are annotated in each panel.}
  \label{fig:preds-multi}
\end{sidewaysfigure}

The model produces per-frame force predictions that closely track
\emph{in vivo} implant recordings across joint types and activities.
Fig.~\ref{fig:preds-multi} shows representative held-out predictions
for two patients evaluated under leave-one-subject-out
cross-validation: a hip case (H3L, top, performing walking with
crutches, aerobics, leg press, and stand-up/sit-down) and a knee case
(K1L, bottom, performing stair ascent, one-leg stance, deep knee
bend, and cycling). Predicted mean forces follow the magnitude, timing, and
shape of the measured forces along all three components, with the
$\pm 2\hat{\sigma}^{\mathrm{cal}}$ bands contracting in
near-stationary phases and widening at peaks and transitions.

\begin{figure}[!htbp]
  \centering
  \includegraphics[width=\linewidth]{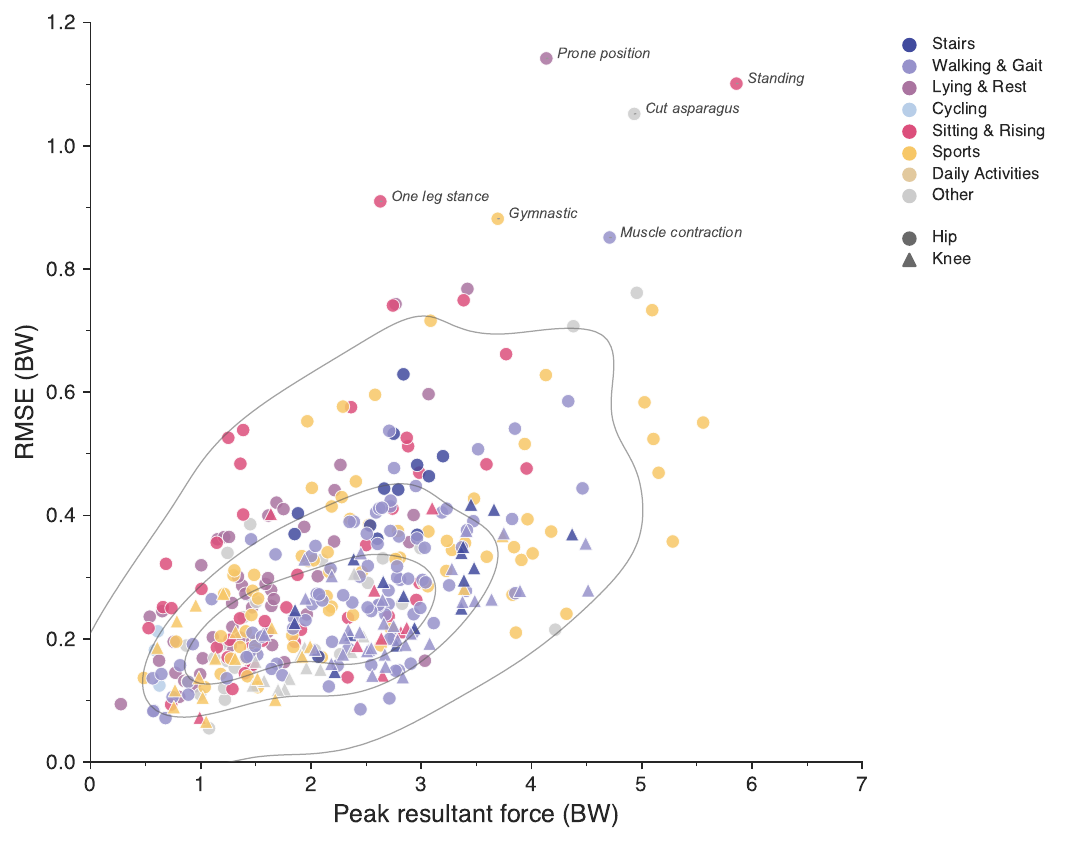}
  \caption{Per-trial prediction error (RMSE) versus peak joint resultant force, both expressed in body weights (BW), evaluated on held-out folds from leave-one-subject-out cross-validation. A stratified subsample of 400 trials is shown for readability. Each point represents one trial, colored by activity category\protect\footnotemark{} and shaped by joint (circle = hip, triangle = knee). Contour lines show the kernel density estimate of the full dataset. Italic labels mark the six outlier trials with the highest RMSE.}
  \label{fig:error-vs-peak-force}
\end{figure}

Fig.~\ref{fig:error-vs-peak-force} plots per-trial RMSE against peak
resultant force across all LOSO folds. The bulk of trials cluster
at low error, with median per-trial RMSE of 0.26\,BW (IQR $[0.19, 0.36]$) across peak forces of $\sim$0.66--4.49\,BW (central 95\% of trials).
A linear mixed-effects model with random
subject and activity effects ($n = 2{,}600$ trials, 26 subjects, 25 activities)
quantifies this scaling: per-trial RMSE grows by 0.090\,BW per BW of
peak resultant force (95\% CI $[0.076, 0.104]$, $p < 0.001$), well
below the diagonal slope of 1 that would correspond to constant
relative error: nRMSE in fact decreases from ${\sim}15\%$
at 1\,BW peak force to ${\sim}11\%$ at 5\,BW. Relative accuracy thus improves with loading magnitude.
The model achieves a median per-implant nRMSE
of 12.4\% (IQR $[9.1, 15.1]$) across the 28 implants. Per-component median
Pearson $r^2$ across LOSO held-out trials is 0.68 for the dominant
axial component ($F_z$) and 0.38/0.29 for the smaller
medial--lateral ($F_x$) and anterior--posterior ($F_y$) components
(per-activity breakdown in Supplementary Fig.~\ref{fig:r2_box}).
Per-cycle peak force $\mathrm{MDC}_{95}$ was 0.20/0.19\,BW
(8.1\%/7.4\% of trial-mean ground-truth peak) for hip/knee during
walking, 0.45/0.50\,BW (17.1\%/14.6\%) during stair ascent, and
0.59/0.45\,BW (17.8\%/11.9\%) during stair descent
(Table~\ref{tab:mdc}).

\begin{table}[!htbp]
  \centering
  \caption{Clinical sensitivity: pipeline minimum detectable change
    ($\mathrm{MDC}_{95}$) versus published effect sizes and MDCs.
    All values are derived from peak resultant forces, in body weights
    (BW). Effect size denotes between-cohort differences
    (e.g.,\ OA vs.\ healthy) or pre/post intervention changes. Stair values
    are reported as ascent\,/\,descent. $\dagger$~Reports
    $\mathrm{MDC}_{90}$ rather than $\mathrm{MDC}_{95}$.}
  \label{tab:mdc}
  \setlength{\tabcolsep}{4pt}
  \begin{tabular}{@{}llccc@{}}
    \toprule
    Clinical comparison                 & Joint                 & Effect size                                                            & Literature                                                                                 & This work               \\
    \midrule
    \multicolumn{5}{@{}l}{\emph{Walking}}                                                                                                                                                                                                                       \\
    \midrule
    OA vs.\ healthy                     & \multirow{3}{*}{Hip}  & ${\sim}0.3$--$0.4$~\citep{diamond2020individuals,van2023biomechanical} & \multirow{3}{*}{$0.34^\dagger$~\citep{cornish2024hip}}                                     & \multirow{3}{*}{$0.20$} \\
    Step length biofeedback             &                       & ${\sim}0.39$~\citep{diamond2022feasibility}                            &                                                                                            &                         \\
    Hip abductor strengthening          &                       & ${\sim}0.72$~\citep{myers2019simulated}                                &                                                                                            &                         \\
    \cmidrule(lr){1-5}
    OA progressors vs.\ non-progressors & \multirow{3}{*}{Knee} & ${\sim}0.4$--$0.6$~\citep{amiri2023high}                               & \multirow{3}{*}{$0.66^\dagger$ / $0.97$~\citep{gardinier2013minimum,price2017reliability}} & \multirow{3}{*}{$0.19$} \\
    Coordination retraining             &                       & ${\sim}0.38$~\citep{uhlrich2022muscle}                                 &                                                                                            &                         \\
    Hip abductor strengthening          &                       & ${\sim}0.42$~\citep{myers2019simulated}                                &                                                                                            &                         \\
    \midrule
    \multicolumn{5}{@{}l}{\emph{Stair negotiation}}                                                                                                                                                                                                             \\
    \midrule
    OA vs.\ healthy                     & Hip                   & $0.5$--$1.7$~\citep{van2023biomechanical}                              & ---                                                                                        & $0.45$ / $0.59$         \\
    OA vs.\ healthy                     & Knee                  & $0.5$--$1.7$~\citep{van2023biomechanical}                              & $1.93$~\citep{price2017reliability}                                                        & $0.50$ / $0.45$         \\
    \bottomrule
  \end{tabular}
\end{table}

\footnotetext{For clarity, activities were grouped into seven broad categories: Stairs (stair ascent/descent, knee brace trials), Walking \& Gait (level walking, gait analysis, footwear conditions, stumbling), Lying \& Rest (bed, lying), Cycling (bicycle), Sitting \& Rising (chair, sitting, deep knee bends, standing), Sports (dance, trampoline, cross-country skiing), and Daily Activities (car, bathing, putting on shoes); remaining activities (muscle contraction and stretching, vibration, agriculture) were pooled as Other.}

\subsection{Generalization across patients and naturalistic activities}
\label{sec:results-generalization}

\begin{figure}[!htbp]
  \centering
  \includegraphics[width=\linewidth]{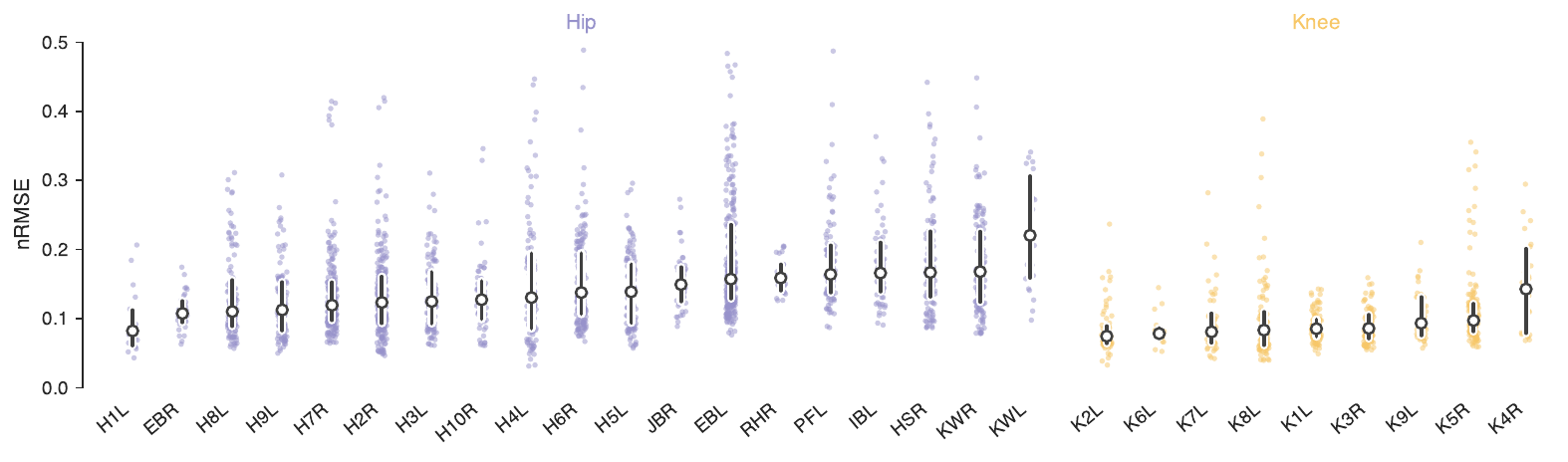}
  \caption{Per-implant prediction error on held-out folds from leave-one-subject-out cross-validation, reported as normalized RMSE (nRMSE). Each column represents one implant, ordered by median nRMSE within joint type. Individual trial errors are shown as jittered points; the vertical bar and dot indicate the interquartile range and median, respectively.}
  \label{fig:per-implant-nrmse}
\end{figure}

To verify that aggregate accuracy is not carried by a small number of
easy cases, performance is disaggregated by implant; predictions for each come from the model that held out its patient (26 folds; both sides of bilateral patients EB and KW held out together). Fig.~\ref{fig:per-implant-nrmse} reports the per-trial nRMSE
distribution for each of the 28 implants. Median nRMSE
ranges from 7.4\% to 22.0\% across the cohort, with
27 of 28 implants (18 of 19 hip, 9 of 9 knee) within
$\pm 5$ percentage points of the median (12.4\%); no single
implant dominates the aggregate. Even the hardest cases (KWL
median 22.0\% for hip, K4R median 14.3\% for knee) show bounded
per-trial IQRs ($[15.9, 30.6]$ and $[7.9, 20.1]$
respectively). By joint type, per-implant median nRMSE was 13.8\% (IQR $[12.1, 16.1]$) at the hip and 8.5\% (IQR $[8.1, 9.3]$) at the knee, with corresponding median RMSEs of 0.27\,BW (IQR $[0.24, 0.32]$) and 0.20\,BW (IQR $[0.18, 0.24]$). For comparison with prior work, the mean $\pm$ SD across implants was $0.32 \pm 0.08$\,BW ($15.9 \pm 3.7\%$) at the hip and $0.23 \pm 0.03$\,BW ($10.2 \pm 2.1\%$) at the knee (Table~\ref{tab:accuracy-comparison}).

\begin{figure}[!htbp]
  \centering
  \includegraphics[width=\linewidth]{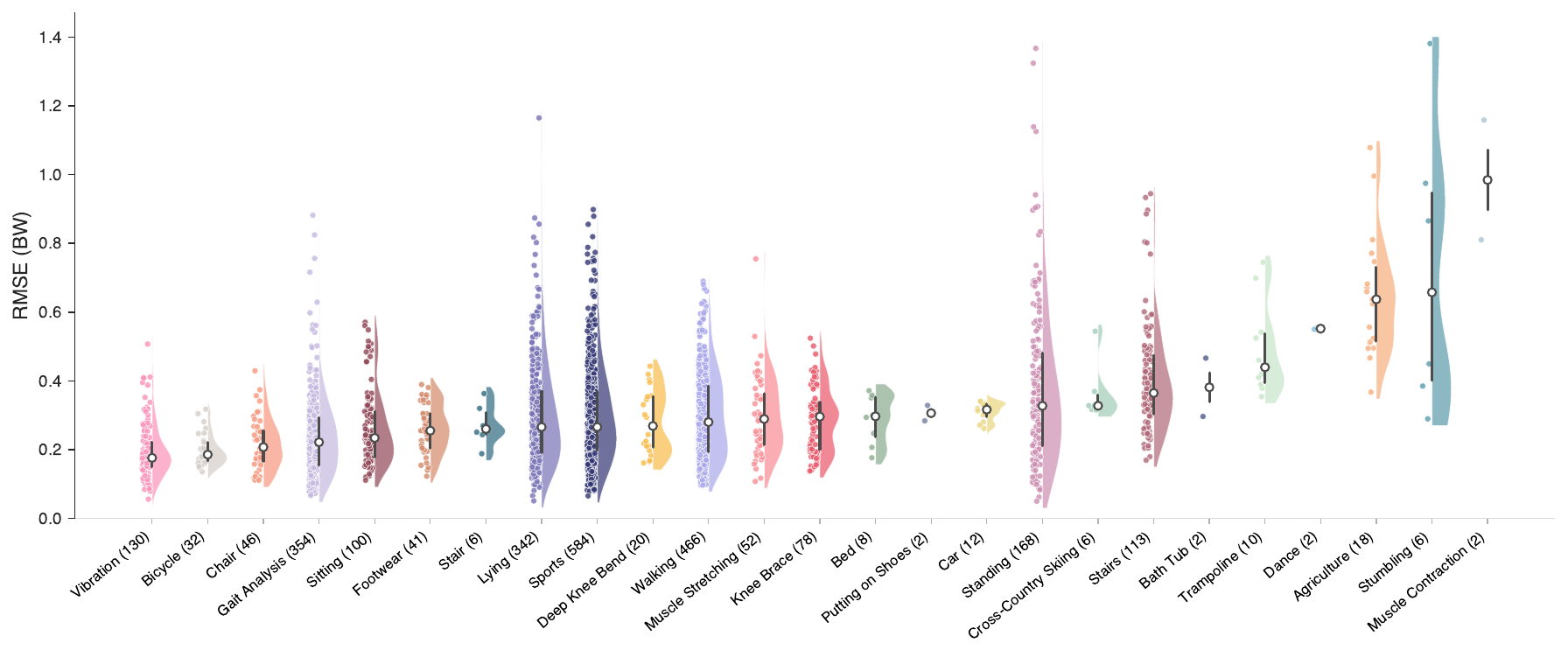}
  \caption{Distribution of per-trial RMSE (in body weights) across activity
    categories, evaluated on held-out folds from leave-one-subject-out
    cross-validation. Each column shows a half-violin
    density estimate, individual trial errors as jittered points, and a summary
    marker indicating the median (dot) and interquartile range (vertical bar).
    Activities are sorted by median RMSE; sample counts are shown in parentheses.}
  \label{fig:raincloud-rmse}
\end{figure}

Performance is broadly stable across the 25 activity categories (Fig.~\ref{fig:raincloud-rmse}). The 14 with at least 20 trials show median RMSE spanning 0.18--0.36\,BW. Five form a
separate high-error cluster with median RMSE between 0.44 and
0.95\,BW: Dance ($n{=}2$), Trampoline ($n{=}10$), Agriculture
($n{=}18$), Stumbling ($n{=}6$), and Muscle Contraction ($n{=}2$);
together 38 trials, 1.5\% of the cohort, all among the
least-represented in training. The full activity~$\times$~implant
cross-tabulation of RMSE and nRMSE (median, IQR) is provided in Supplementary
Tables~\ref{tab:rmse-hip-1}--\ref{tab:rmse-hip-2} (hip) and
Table~\ref{tab:rmse-knee} (knee).

\begin{table}[t]
  \centering
  \caption{Comparison with published joint contact force estimation
    methods. All errors are RMSE in body weights (BW) or normalized RMSE
    (\%). Values for this work are mean $\pm$ SD across implants; figures for prior methods are reproduced as reported in the cited sources. This work covers 25 activity categories, whereas all
    listed comparison methods are restricted to gait or a small set of
    prescribed tasks.
    Methods in the bottom section use fully out-of-laboratory inputs
    (i.e., no optical markers, force plates, or multi-channel EMG).
    GRF, ground reaction force; EMG, electromyography; CT, computed
    tomography; IMU, inertial measurement unit; fluoro, fluoroscopy.
    $\dagger$~Validated against \emph{in silico} musculoskeletal
    estimates rather than \emph{in vivo} instrumented implants.
    $\ddagger$~Evaluated on 6 of 9 knee patients from this study.
    $\S$~Medial compartment force only; total contact force RMSE would
    be higher. $\|$~Amiri: walking, stairs, sit/stand; Derungs: walking, squatting, stairs, sit/stand; Peng: walking, running. All unmarked methods evaluate walking only.}
  \label{tab:accuracy-comparison}
  \small
  \begin{tabular}{@{}lllll@{}}
    \toprule
    Method                                    & Joint      & Input                        & RMSE (BW)       & nRMSE (\%)     \\
    \midrule
    This work (LOSO)                          & Hip        & Monocular video              & $0.32 \pm 0.08$ & $15.9 \pm 3.7$ \\
    This work (LOSO)                          & Knee       & Monocular video              & $0.23 \pm 0.03$ & $10.2 \pm 2.1$ \\
    \midrule
    \citet{amiri2022prediction}$^\|$          & Hip        & Markers + GRF                & $0.17$--$0.60$  & ---            \\
    \citet{cornish2024hip}$^\dagger$          & Hip        & Markers + EMG                & $0.47 \pm 0.24$ & $13.4 \pm 7.1$ \\
    \citet{princelle2025emg}                  & Knee       & Markers + GRF + EMG + CT     & $<0.56$         & ---            \\
    \citet{rabbi2024muscle}$^\dagger$         & Knee       & Markers + GRF + EMG          & $0.19 \pm 0.05$ & ---            \\
    \citet{zou2024prediction}$^\S$            & Knee       & Markers + GRF + EMG          & $0.21$--$0.38$  & ---            \\
    \citet{derungs2026machine}$^{\ddagger\|}$ & Knee       & Markers + fluoro + GRF + EMG & ---             & $11.9$--$23.4$ \\
    \midrule
    \citet{diraimondo2023peak}$^\dagger$      & Knee       & IMU                          & $0.40 \pm 0.17$ & ---            \\
    \citet{peng2024smartphone}$^{\dagger\|}$  & Hip / Knee & Stereo video                 & $0.23$--$0.77$  & ---            \\
    \bottomrule
  \end{tabular}
\end{table}

\subsection{Generalization to an independent instrumented cohort}
\label{sec:grand-challenge}

On the six Grand Challenge competitions (Sec.~\ref{sec:methods-grand-challenge}; 195 trials from 4 instrumented patients), per-trial RMSE was 0.45\,BW (IQR $[0.39, 0.53]$) with median $r^2$ of 0.81 (IQR $[0.74, 0.85]$) (Fig.~\ref{fig:error-vs-peak-force-fregly}, Table~\ref{tab:per-trial-resultant}). This is more than double the in-distribution LOSO knee RMSE of 0.20\,BW.
Per-competition median RMSE spanned 0.41--0.56\,BW (highest for GC~2), while the largest individual errors were crouch and bouncy-gait trials from GC~3, all exceeding 0.9\,BW. Fig.~\ref{fig:gc-percentile-traces} shows
representative traces sampled across the RMSE distribution.
Table~\ref{tab:gc-comparison} summarizes the per-competition comparison
with published winning entries.

\begin{table}[t]
  \centering
  \caption{Per-competition comparison with Grand Challenge
    winners~\citep{fregly2012grand}. RMSE in body weights (BW),
    averaged over the competition's evaluated trials.
    Winner results are taken from the original publications.
    $\natural$~Evaluated on overground walking trials only, as reported by \citet{thelen2014co}.}
  \label{tab:gc-comparison}
  \begin{tabular}{@{}clcc@{}}
    \toprule
    GC                                          & Winner                       & Winner (BW)                   & This work (BW)                  \\
    \midrule
    1                                           & \citet{kim2010effect}        & 0.61 / 0.72                   & \textbf{0.51} /   \textbf{0.49} \\
    2                                           & \citet{hast2013dual}         & \textbf{0.51} / 0.81          & 0.53 / \textbf{0.66}            \\
    3                                           & \citet{manal2012predictions} & \textbf{0.35} / 0.79          & 0.51 / \textbf{0.69}            \\
    3                                           & \citet{knowlton2012grand}    & \textbf{0.34} / \textbf{0.63} & 0.51 / 0.69                     \\
    4\rlap{$^{\natural}$}                       & \citet{thelen2014co}         & 0.51                          & \textbf{0.38}                   \\
    5                                           & \citet{marra2015subject}     & \textbf{$<$0.30} / $<$0.40    & 0.43 / \textbf{0.32}            \\
    6                                           & \citet{jung2016intra}        & 0.51 / 0.42                   & \textbf{0.44} / \textbf{0.32}   \\
    \midrule
    \multicolumn{2}{@{}l}{Overall (195 trials)} &                              & $0.45$ [IQR $0.39$--$0.53$]                                     \\
    \bottomrule
  \end{tabular}
\end{table}

\subsection{Input-dependent and calibrated predictive uncertainty}

The heteroscedastic head produces uncertainty estimates whose temporal
modulation tracks task dynamics. Across the LOSO held-out cohort
($n = 2{,}600$ trials), the within-trial coefficient of variation
$\sigma_{\mathrm{std}}/\sigma_{\mathrm{mean}}$ on the dominant axial
component is highest for cyclic, weight-bearing tasks
(median $F_z$ $\sigma$-CV $= 0.20$, IQR $[0.14, 0.24]$;
e.g.\ Stairs 0.25, Walking 0.20, Footwear 0.21) and lowest for
quasi-static activities
(median 0.07, IQR $[0.03, 0.12]$;
e.g.\ Vibration 0.02, Lying 0.06, Bicycle 0.08), a ${\sim}3\times$
contrast (Fig.~\ref{fig:sigma_cv}). For tasks with little temporal
structure in the target signal, $\hat{\sigma}$ remains near-uniform;
for cyclic tasks, $\hat{\sigma}$ widens at peaks and
transitions and tightens during stable phases, indicating that
predicted uncertainty responds to input dynamics rather than to a
global noise floor.

After per-axis temperature scaling (Methods,
Sec.~\ref{sec:methods_calibration}), the predictive bands are
well-calibrated in absolute terms: empirical coverage at
$\pm 2\hat{\sigma}^{\mathrm{cal}}$ is 92.5\%, 94.7\%, and 95.4\%
for $F_x$, $F_y$, and $F_z$ respectively, within 3 percentage points
of the nominal 95.45\% expected under a Gaussian likelihood.
The fitted temperatures
($\tau_x = 1.87$, $\tau_y = 1.67$, $\tau_z = 1.97$)
are highly stable across LOSO folds (per-fold standard deviation
$\leq 0.04$). Because temperature scaling is multiplicative, it
preserves the heteroscedastic structure characterized in
Fig.~\ref{fig:sigma_cv}; the calibrated
$\pm 2\hat{\sigma}^{\mathrm{cal}}$ bands are therefore well-approximated
as 95\%-credible intervals on the per-frame force prediction.
Stratified by peak-$|F_z|$ tercile, empirical coverage is 97.7\%
in the lowest tercile, 98.4\% in the middle, and 87.5\% in the
highest, indicating mild over-conservatism across non-peak loading
and mild over-confidence at peaks.

\subsection{Self-supervised video features substitute for activity labels}

The contribution of each input modality is quantified next.
Table~\ref{tab:ablation} reports nRMSE for progressively richer input
configurations on the held-out 85/15 patient-stratified validation
split. Kinematics alone yield 16.8\% overall nRMSE. Adding the SMPL
shape vector $\bm{\beta}$ produces a marginal improvement
($-0.5$ percentage points). Adding the activity text embedding
reduces nRMSE to 13.5\% ($-2.8$ pp), the largest single auxiliary
contribution. \vjepatwo{} features yield a further improvement to
12.8\% ($-0.7$ pp), with the largest gains on knee predictions
($9.8\% \to 8.9\%$).
The \vjepatwo{}-without-text variant matches the full text-and-video
model (12.9\% versus 12.8\%) despite never seeing the curated
activity label.

Both auxiliary modalities substantially reduced nRMSE relative to the kinematics + shape baseline (text: 13.5\% versus 16.3\%; \vjepatwo{}: 12.9\%), with
\vjepatwo{} features yielding a small additional gain of 0.6 percentage points over text embeddings. Per-trial improvements from the two modalities were highly
correlated across the validation set ($r = 0.86$, $p < 0.001$; Fig.~\ref{fig:vjepa_breakdown}C), indicating that text and \vjepatwo{} features capture
a largely-shared activity-related signal at the per-trial level. At the category level,
three activities of the 14 categories tested had CIs favoring \vjepatwo{} (aerobics, deep knee bend, and stair negotiation) and none favored text (Fig.~\ref{fig:vjepa_breakdown}B). Together, these results suggest that self-supervised video representations can substitute for curated activity annotations without loss of accuracy, eliminating a manual labeling bottleneck for clinical deployment.

\begin{table}[!htbp]
  \centering
  \caption{Ablation study on input modalities. Normalized RMSE
    (nRMSE) is reported on the validation set.
    K: kinematics, S: shape parameters, T: text embeddings, V: \vjepatwo{}
    video features.}
  \label{tab:ablation}
  \begin{tabular}{lccccccc}
    \toprule
                         & \multicolumn{4}{c}{Inputs} & \multicolumn{3}{c}{Val nRMSE(\%)}                                          \\
    \cmidrule(lr){2-5} \cmidrule(lr){6-8}
    Configuration        & K                          & S                                 & T      & V      & Overall & Hip & Knee \\
    \midrule
    Kinematics only      & \cmark                     &                                   &        &        & 16.8    &
    19.3                 & 11.9                                                                                                    \\
    + Shape              & \cmark                     & \cmark                            &        &        & 16.3    &
    18.8                 & 11.5                                                                                                    \\
    + Text               & \cmark                     & \cmark                            & \cmark &        & 13.5    &
    15.5                 & 9.8                                                                                                     \\
    + \vjepatwo{}        & \cmark                     & \cmark                            & \cmark & \cmark &
    \textbf{12.8}        & 14.8                       & \textbf{8.9}                                                               \\
    \vjepatwo{}, no text & \cmark                     & \cmark                            &        & \cmark & 12.9    &
    \textbf{14.6}        & 9.5                                                                                                     \\
    \bottomrule
  \end{tabular}
\end{table}

\subsection{Closed-loop motion redesign reduces joint loading}

With prediction quality established across patients, activities, and
loading regimes, the trained predictor exposes a differentiable
surrogate from kinematics to joint forces, the natural target for
gradient-based motion design. This surrogate is combined with the
rectified-flow motion prior described in Sec.~\ref{sec:inverse_design},
and SDEdit-style guided generation is steered toward reduced peak axial
loading.

\begin{figure}[!htbp]
  \centering
  \includegraphics[width=\linewidth]{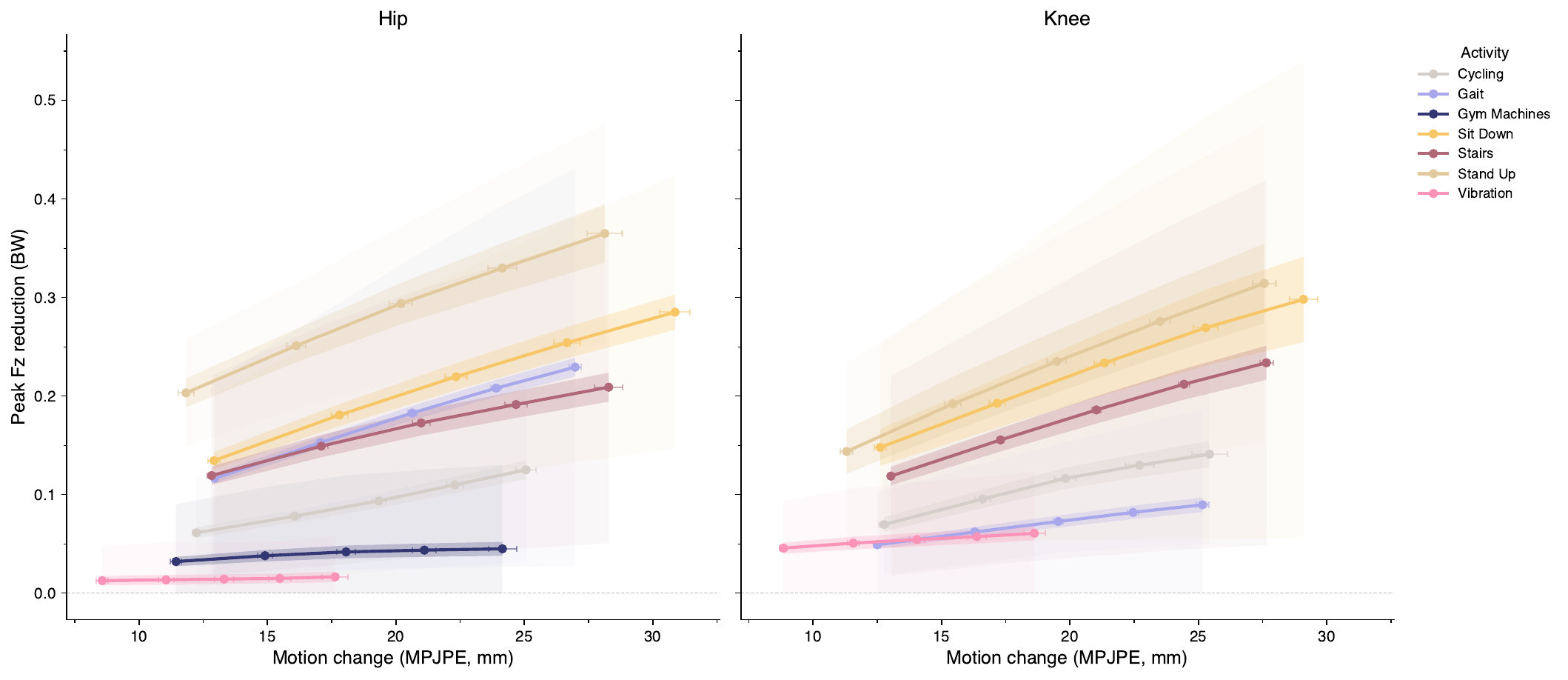}
  \caption{Activity modifiability: force reduction versus motion change at the hip (left) and knee (right). The horizontal axis shows the mean per-joint position error (MPJPE) between the original and optimized motion, quantifying the magnitude of kinematic modification, while the vertical axis shows the resulting reduction in peak axial joint contact force ($F_z$, in body-weight units). Each colored curve represents an activity category, aggregated across all trials in that category, with markers corresponding to increasing noise levels ($t_\text{start} \in \{0.10, 0.15, 0.20, 0.25, 0.30\}$) and horizontal whiskers showing the standard error of the per-category mean MPJPE. The darker inner band denotes $\pm$1 standard error of the category mean (uncertainty about the average behavior, which shrinks with sample size), while the wider outer band denotes $\pm$1 standard deviation across trials within the category (the inherent spread of trial outcomes, independent of sample size). Steeper curves indicate activities where small motion adjustments yield large force reductions; these represent the highest-value targets for clinical motion retraining. Modifiability varies markedly across activities, from large reductions in sit-to-stand and stair negotiation to little or no change in motion imposed by external apparatus.}
  \label{fig:modif-curves}
\end{figure}

\paragraph{Force reductions across activities.}
Fig.~\ref{fig:modif-curves} plots, for each held-out trial, the
reduction in peak $F_z$ against the mean per-joint position error
(MPJPE) between the original and optimized motion, swept over the
SDEdit start time $t_\text{start} \in \{0.10, 0.15, 0.20, 0.25, 0.30\}$.
Across all trials, guided generation at $t_\text{start} = 0.30$ reduced the predicted peak $F_z$ by a median
of 0.12\,BW (IQR $[0.05, 0.24]$) at a median MPJPE of
26\,mm. Activities involving dynamic weight
transfer (sit-to-stand and stair negotiation) produce the steepest
curves (mean reductions of 0.24 and 0.22\,BW respectively), indicating
that small kinematic adjustments suffice for substantial unloading.
Static or constrained activities (gym machines,
vibration plates) yield the lowest, flattest curves (median reductions all below
0.05\,BW): the original motion already operates near a local minimum of
the predicted load, leaving little room for guided modification.

\begin{figure}[!htbp]
  \centering
  \includegraphics[width=\linewidth,height=0.75\textheight,keepaspectratio]{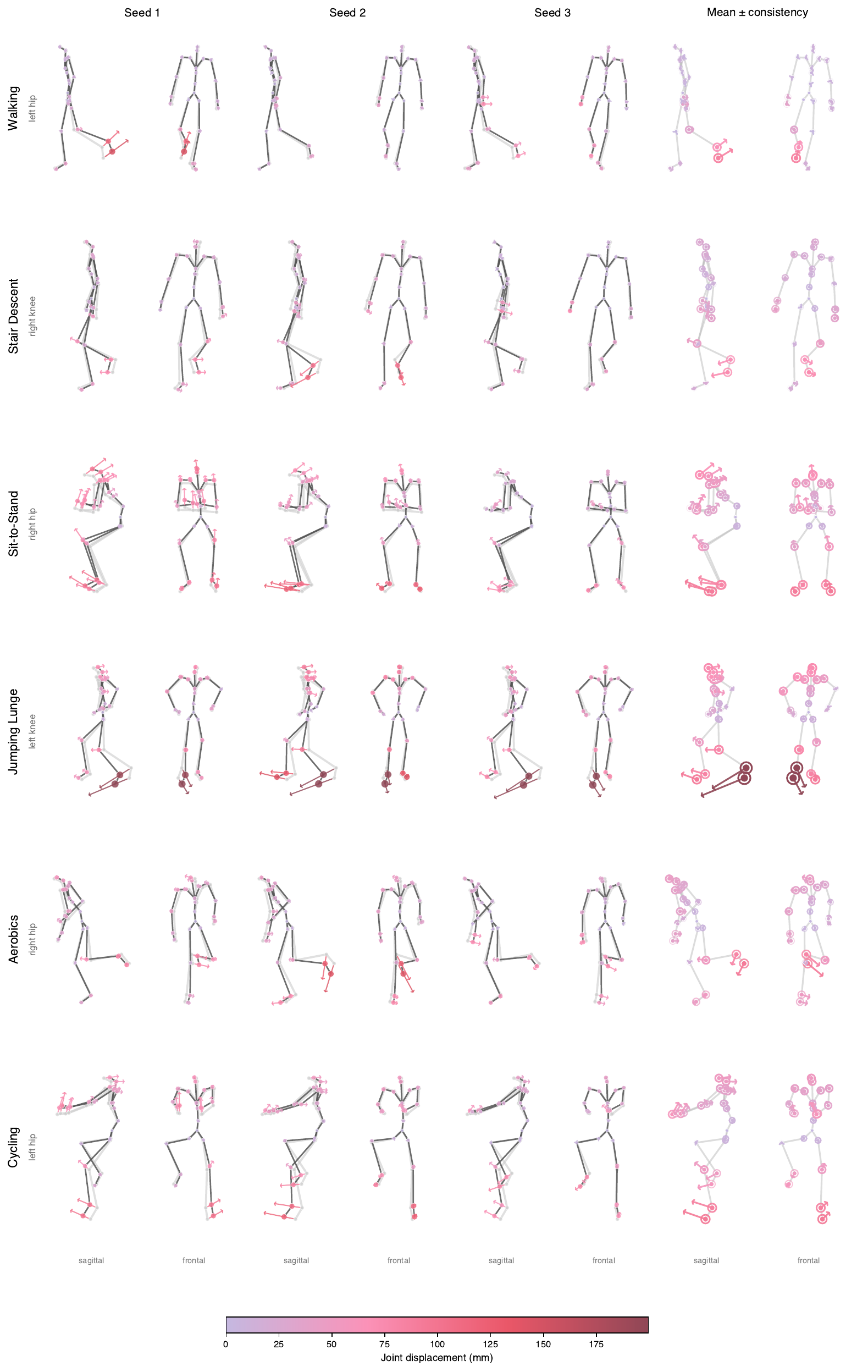}
  \caption{Motion strategy characterization at peak force frame. Each row shows a representative activity; columns display three independent optimization runs (Seeds 1--3) and the seed-averaged displacement (rightmost). For each run, the original pose is shown in light gray and the optimized pose in dark gray, with joints colored by displacement magnitude (lavender--berry colormap). Arrows indicate the direction and relative magnitude of joint displacement ($3\times$ amplified for visibility). Sagittal and frontal views are shown side by side for each condition. In the mean column, concentric rings encode cross-seed directional consistency: large, opaque rings indicate that all seeds displaced that joint in the same direction, suggesting a robust biomechanical strategy rather than an optimization artifact. Per-activity strategies: \emph{walking}---increased contralateral knee flexion during early swing; \emph{stair descent}---trailing leg kept closer underneath the pelvis, with slight lateral lean toward the lead-foot side; \emph{sit-to-stand}---feet positioned beneath the knees with a more upright trunk; \emph{jumping lunge}---trailing limb closer to the lead leg (reduced hip extension and adduction) with a more upright trunk; \emph{aerobics}---reduced contralateral knee flexion and hip abduction; \emph{cycling}---anterior shift of the lower limb relative to the pelvis (analogous to increased saddle setback) combined with elevated handlebar height.}
  \label{fig:strategy-sticks}
\end{figure}

\paragraph{Optimized strategies are consistent across seeds and biomechanically interpretable.}
Three independent optimization seeds were run for each activity
(Fig.~\ref{fig:strategy-sticks}), with joints colored and arrowed by
their per-joint displacement at the peak force frame; the rightmost
column overlays cross-seed displacements as concentric rings whose
opacity encodes directional agreement. For example, when walking, the contralateral foot is displaced such that the knee is in greater flexion during early swing; during a sit-to-stand, feet are displaced under the knees and trunk flexion is reduced.

\begin{sidewaysfigure}
  \centering
  \includegraphics[width=\textheight]{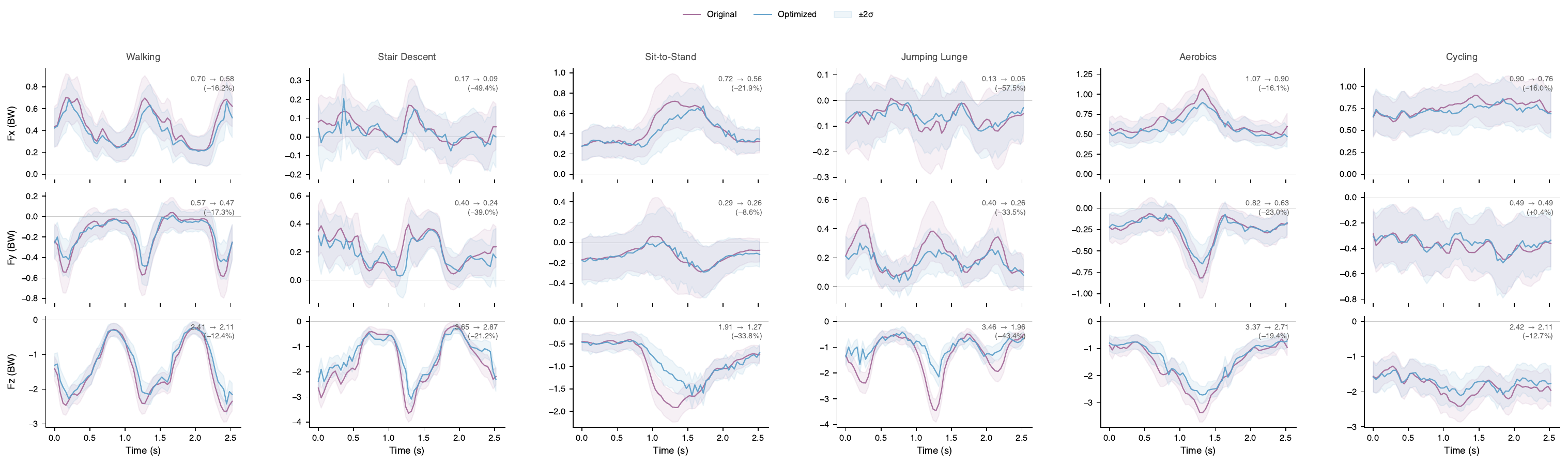}
  \caption{Force time series before and after motion optimization across
    six representative activities. Each column corresponds to a different
    activity; rows show the three force components ($F_x$, $F_y$, $F_z$)
    in units of body weight (BW). Original forces are shown in mauve and
    optimized forces in blue, with shaded bands indicating $\pm 2\sigma$
    prediction uncertainty. Per-component peak
    force changes are annotated in each panel. The optimization targets
    peak axial force ($F_z$) reduction; the rectified-flow prior
    constrains the motion to remain on-manifold (kinematic visualizations
    in Fig.~\ref{fig:strategy-sticks}).}
  \label{fig:force-comparison}
\end{sidewaysfigure}

\paragraph{Plausible kinematic changes yield meaningful force reductions.}
The kinematic strategies in Fig.~\ref{fig:strategy-sticks} translate
into the force trajectories of Fig.~\ref{fig:force-comparison}, which
compares pre- and post-optimization 3D force time series for six
representative activities. Peak $F_z$ decreases by 0.04--0.60\,BW
(median 9\%). Across
all trials, the predictor's per-axis standard deviation on
the optimized motion was within $\pm 4\%$ of its value on the
original motion (median $\bar{\sigma}_\text{opt}/\bar{\sigma}_\text{orig}$:
0.98, 0.98 and 0.99 for $F_x$, $F_y$ and $F_z$, respectively; IQR $[0.96, 1.00]$); no trial
exhibited $\bar{\sigma}_\text{opt}/\bar{\sigma}_\text{orig} > 1.5$ on
any axis.

\section{Discussion}

I trained a single-camera, physics-free pipeline for predicting hip
and knee contact forces with calibrated uncertainty, validated
against \emph{in vivo} implant recordings across 26 patients and 25 activities. Prediction accuracy matches that of laboratory musculoskeletal
pipelines that require far richer instrumentation, and the trained model's gradients are biomechanically meaningful: they guide a generative motion prior toward load-reducing strategies that align with the biomechanics literature.

\subsection{Laboratory-grade joint contact force accuracy without musculoskeletal modeling}

Unlike physics-based pipelines---which depend on assumptions about
muscle recruitment, joint center definitions, soft tissue artifact
corrections, Hill-type contractile dynamics, and contact
mechanics---this pipeline learns a single end-to-end mapping from
direct measurements at the prosthesis. Per-frame uncertainty bands
emerge from the same network and are well-calibrated in aggregate
(with mild conditional departures detailed in Sec.~\ref{sec:limitations}), a
property biomechanical simulation does not natively produce. They let downstream users weight
individual predictions by their reliability and support
uncertainty-aware optimization in the inverse design pipeline.

Musculoskeletal pipelines are mature but not robust to their own
assumptions: muscle--tendon parameter
uncertainty alone can swing predicted knee forces by up to 2.1\,BW
during a bodyweight squat~\citep{hosseini2022uncertainty}, and
systematic reviews have reached no consensus on how model choices
affect hip and knee force predictions~\citep{moissenet2017alterations}.
Against this baseline, the present approach matches these laboratory methods in absolute
terms despite requiring only monocular video
(Table~\ref{tab:accuracy-comparison}), with tighter dispersion across a far broader activity repertoire than any single
comparison method. On 6 of the 9 knee patients evaluated here,
\citet{derungs2026machine} report nRMSE between $11.9\%$ and $23.4\%$
across walking, squatting, stairs, and sit/stand using markers,
fluoroscopy, ground reaction forces, and surface EMG; the current video-only
model achieves $6.8\%$--$9.4\%$ on the same patients and analogous
activities (mean across per-subject means, their convention).
Notably, much of the broader literature validates against
\emph{in silico} forces produced by a
musculoskeletal simulation rather than \emph{in vivo} implant
recordings: a fundamental ceiling on attainable accuracy that the
present pipeline avoids by training end-to-end on direct measurements
at the prosthesis.

No published \emph{in vivo} evaluation of joint contact force estimation has tested across cohorts: existing studies train and test within the same instrumented patients, conflating model capacity with cohort-specific calibration. Applied without retraining to the Grand Challenge datasets~\citep{fregly2012grand}---a separate cohort with cruciate-retaining rather than cruciate-sacrificing prostheses---the method improves on published winners in 3 of 6 competitions and rivals them on the remainder (Table~\ref{tab:gc-comparison}). Against the recent CT-personalized and EMG-informed neuromusculoskeletal pipeline of \citet{princelle2025emg}, predictions are comparable on two of four subjects and behind on two. The ${\sim}0.25$\,BW gap to LOSO plausibly reflects three concurrent shifts: implant and transducer architecture, an older cohort performing prescribed gait modifications absent from training, and deliberately conservative inference (bone-frame predictions against implant-frame ground truth, \vjepatwo{} pathway zeroed). Errors concentrate on the perturbed gaits (crouch, bouncy) rather than baseline walking, pointing to cohort and protocol as the dominant factor.

This substitution, however, is not assumption-free. The musculoskeletal pipeline
assumes a model and degrades gracefully outside its calibrated range
because its mechanical structure remains valid; the data-driven
pipeline instead trades model fidelity for generalization, with
out-of-distribution failures that are correspondingly less
interpretable. The empirical
question is which set of assumptions is cheaper to satisfy at scale.
For functional activities within OrthoLoad's coverage, the results above
show the learned mapping is equally accurate at a fraction of the cost:
inference takes under a minute per trial on a consumer GPU, with no
subject-specific calibration. The approach is also permissive
about acquisition: the camera may be uncalibrated and moving, with no
pose refinement, foot--floor contact handling, or ground plane
estimation of the kind physics-based alternatives require, bringing head-mounted wearables such as smart glasses within reach as capture devices.

Clinically, the model's $\mathrm{MDC}_{95}$ during
walking is substantially
tighter than published values for calibrated musculoskeletal
simulations and resolves the principal effect sizes in the gait
retraining, strengthening, and osteoarthritis literature; sensitivity degrades during stair negotiation but remains within clinically relevant cohort separations (Table~\ref{tab:mdc}). This precision appears sufficient
to track per-patient response to intervention and stratify cohorts by
baseline loading, applications previously confined to instrumented
laboratories.

\subsection{Self-supervised video features as activity context priors}

Curated activity strings---expensive to collect, brittle in
deployment, and dependent on a controlled vocabulary that may not
transfer outside research settings---are unexpectedly dispensable.
A frozen \vjepatwo{}~\citep{assran2025vjepa} feature stream alone cuts overall
validation nRMSE from 16.3\% (kinematics + shape baseline) to
12.9\%, a 3.4-percentage-point absolute reduction that matches
the 12.8\% obtained when text and \vjepatwo{} are supplied together: the curated label adds nothing once video features are available. In aggregate the two signals contribute almost equally, but \vjepatwo{} holds a per-category edge precisely where text labels most underdetermine execution, making the
curated label the more easily eliminated of the two.

That \vjepatwo{} features can substitute for explicit text labels is grounded in the model's architectural priors and pretraining objective. Unlike generative video models that optimize
for pixel-level reconstruction (expending capacity on unpredictable
high-frequency details), \vjepatwo{} employs a joint-embedding
predictive architecture that operates entirely in a learned
representation space~\citep{assran2025vjepa}; predicting the latent
representations of masked spatiotemporal patches forces the encoder
to capture predictable underlying dynamics rather than appearance
cues, yielding strong performance on benchmarks deliberately
constructed so that single-frame appearance is insufficient. The resulting video representation acts as a dense, continuous activity label: while a discrete text string like ``stair ascent with handrail'' provides a
coarse categorical prior, \vjepatwo{} captures both the semantic
identity of the activity and the fine-grained spatiotemporal
nuances of its execution. Its state-of-the-art performance on action
anticipation~\citep{assran2025vjepa} further suggests that the learned
representation supports linear prediction of the short-horizon
kinematic continuations that activities imply, precisely the signal
a force regressor needs.

Two further observations sharpen the picture. First, adding the SMPL shape vector
$\bm{\beta}$ on top of kinematics yields only a marginal $-0.5$
percentage-point improvement, suggesting that subject morphology
contributes little beyond what segment-relative joint positions
already encode implicitly. Second, pose alone is informative, but
coarse activity context (whether supplied as a text label or
extracted from raw video by a self-supervised encoder) substantially
disambiguates kinematically similar motions whose force profiles
diverge, such as walking versus walking with crutches, or stair ascent with
versus without handrail support.

More broadly, the result suggests that pretrained video world models
may serve as general-purpose context priors for biomechanical
inference. Estimating ground reaction forces, joint moments, and muscle forces
all share the same need for activity disambiguation that \vjepatwo{}
appears to satisfy here without task-specific supervision; whether self-supervised video pretraining
at scale can absorb the role traditionally played by curated metadata
in clinical biomechanics warrants further investigation.

\subsection{Generative inverse design as a hypothesis generator for clinical biomechanics}

Closing the loop from prediction to design unlocks a class of
applications that physics-based pipelines have historically supported
only at high cost. Optimizing a motion through traditional predictive simulation
requires solving a large nonlinear program that simultaneously enforces multibody equations of motion, muscle dynamics, and contact mechanics; here, the
same objective reduces to backpropagation through a single learned and
calibrated network. The combination of a differentiable surrogate
from kinematics to forces with a generative motion prior trained on
the same kinematic distribution defines a fully end-to-end
inverse design pipeline that can be steered toward any differentiable
force objective.

A central concern in any gradient-guided generation procedure is whether
observed force reductions reflect genuine biomechanical strategies or
adversarial exploits of the predictor's gradient field. Cross-seed
consistency rules out the simplest such artifact: across three
independent optimization seeds, joint displacements at the peak force
frame agree in direction for the dominant degrees of freedom, so the
recovered strategies are stable attractors of the optimization rather
than artifacts of any single initialization. Consistency alone, however,
would not separate a genuine strategy from a systematic exploit; what
does is that the emergent kinematics are interpretable in light of
biomechanical reasoning about moment arms, load redistribution, and
muscle demand, and converge on strategies independently established in
the biomechanics literature, which a gradient-field exploit would have no
reason to reproduce. For sit-to-stand specifically, the more upright
trunk found here matches the strategy that a predictive simulation
framework independently identified as a posture that reduces hip
load~\citep{van2024planar}. Likewise during stair descent, retaining the
trailing leg closer underneath the pelvis prolongs trailing limb
weight-bearing, an adjustment known to attenuate impact loads on the
leading knee~\citep{karamanidis2011altered}. For cycling, the resulting
reduction in hip loading corresponds kinematically to greater saddle
setback, which reduces rectus femoris activation~\citep{bini2014effects}
and the hip flexor demand it generates. That post-optimization predictive
uncertainty remains within $\pm 4\%$ of the original on each axis further
indicates that the optimized motions stay inside the model's confident
region rather than drifting into out-of-distribution territory where its
gradients would be unreliable.

Sit-to-stand and stair negotiation produce the steepest
force-versus-MPJPE curves, with mean peak-$F_z$ reductions of 0.24
and 0.22\,BW at modest kinematic perturbation. Apparatus-imposed
motion (e.g., gym machines, vibration plates) yields low, flat curves: the original motion already operates near a local minimum
of the predicted load, so guided generation finds little to modify. Clinically, this is the more useful pattern: the activities
where small kinematic adjustments yield large force reductions are
precisely the ones for which retraining interventions
exist---transfer training, gait modification, stair negotiation
strategies---making the inverse design output a natural input to
motion retraining workflows.

What remains untested is biomechanical translation. The strategies
the model surfaces are predictions about kinematic changes that would
reduce loading according to the learned mapping; whether subjects can
adopt those strategies, whether they remain effective under the
subject's actual physiology rather than the predictor's distillation
of it, and whether the loading reductions persist after the
kinematic perturbation propagates through real muscle actuation are
open questions. The
contribution here is a hypothesis-generating procedure---near-instantaneous,
differentiable, and validated against \emph{in vivo}
measurements---that can prioritize candidate motion modifications for
clinical investigation rather than prescribe them.

\subsection{Limitations}
\label{sec:limitations}

Three limitations bound the claims above. First, the cohort that
supplies ground truth is biased by construction. Instrumented
prostheses are implanted only in arthroplasty patients: typically
elderly, with end-stage joint degeneration and prominent
peri-articular muscle atrophy~\citep{mizner2005early} that often persists
for years post-operatively~\citep{konig2000balance}, and ethically
restricted from vigorous athletic movement. Their contact loads
are thus unlikely to be fully representative of the native
joints of younger, more active populations. The pipeline does transfer zero-shot to the only independent instrumented cohort available (Sec.~\ref{sec:grand-challenge}), but that cohort shares the same profile. Because in vivo ground truth exists only in instrumented patients, accuracy beyond this profile cannot be
established with current data. Second,
stratified calibration analysis reveals an asymmetry by loading
magnitude: the calibrated $\pm 2\hat{\sigma}^{\mathrm{cal}}$ bands
are slightly over-conservative across non-peak loading and slightly
over-confident at peaks. They therefore convey reliable confidence
statements at moderate-magnitude loads but should be read with mild
caution at the peaks, the frames typically of greatest clinical
interest. Third, the model's accuracy is conditioned on the
quality of the upstream monocular 3D mesh recovery: pose estimation
failures propagate to force prediction failures, and the reported
numbers reflect this specific pose stack rather than an
architecture-invariant performance ceiling. The framework is modular
in this regard: the pose estimator can be upgraded as the field
advances, without redesigning the force predictor.

\subsection{Outlook}

The most immediate applications lie where instrumented measurement has never been feasible and recording conditions cannot be controlled: retrospective analysis of archived clinical videos from uncalibrated cameras, rapid screening in primary care before referral, and longitudinal at-home monitoring during
rehabilitation. A companion web interface, offering cloud-based inference, is under development; streaming inference on portable hardware for real-time biofeedback is the next engineering direction.

\section*{Acknowledgments}

I am grateful to Shaokai Ye for discussions on multimodal training, and to Rajat Thomas for a careful reading and thoughtful pushback on generalizability.

  {
    \small
    \bibliographystyle{unsrtnat}
    \bibliography{refs}
  }

\clearpage
\appendix
\renewcommand{\thefigure}{A\arabic{figure}}
\renewcommand{\theHfigure}{A\arabic{figure}}
\setcounter{figure}{0}
\renewcommand{\thetable}{A\arabic{table}}
\renewcommand{\theHtable}{A\arabic{table}}
\setcounter{table}{0}
\section{Supplementary Material}

This supplementary material provides per-activity, per-implant breakdowns of prediction error, along with per-category analyses of temporal shape agreement, predictive uncertainty, and the added value of self-supervised video features over curated activity labels.

\begin{sidewaystable}
  \centering
  \scriptsize
  \setlength{\tabcolsep}{3pt}
  \renewcommand{\arraystretch}{1.1}
  \caption{Hip cohort (part 1/2): per-activity, per-implant prediction error.
    Each cell shows RMSE median (Q1--Q3) in BW (top) with nRMSE median (Q1--Q3)
    in \% on the gray line below. The ``All'' row and column report marginal
    medians over the pooled trial-level distribution across all 19 implants. $^\dagger$Based on fewer than 3 trials.}
  \label{tab:rmse-hip-1}
  \begin{tabular}{lccccccccccc}
\toprule
 & \textbf{EBL} & \textbf{EBR} & \textbf{H1L} & \textbf{H2R} & \textbf{H3L} & \textbf{H4L} & \textbf{H5L} & \textbf{H6R} & \textbf{H7R} & \textbf{H8L} & \textbf{All} \\
\midrule
Vibration & -- & -- & -- & \makecell{0.15\,{\tiny (0.10--0.18)} \\ \textcolor{gray}{10.3\,{\tiny (9.4--12.0)}}} & \makecell{0.19\,{\tiny (0.16--0.27)} \\ \textcolor{gray}{13.7\,{\tiny (12.5--16.0)}}} & \makecell{0.19\,{\tiny (0.18--0.24)} \\ \textcolor{gray}{20.2\,{\tiny (17.5--22.0)}}} & \makecell{0.20\,{\tiny (0.17--0.25)} \\ \textcolor{gray}{15.3\,{\tiny (12.4--17.2)}}} & -- & -- & -- & \makecell{0.18\,{\tiny (0.16--0.23)} \\ \textcolor{gray}{14.6\,{\tiny (11.8--18.1)}}} \\
Bicycle & \makecell{0.22\,{\tiny (0.19--0.24)} \\ \textcolor{gray}{32.0\,{\tiny (27.6--35.3)}}} & -- & -- & -- & -- & -- & -- & -- & -- & -- & \makecell{0.19\,{\tiny (0.17--0.22)} \\ \textcolor{gray}{24.1\,{\tiny (19.3--29.2)}}} \\
Gait Analysis & -- & -- & \makecell{0.19\,{\tiny (0.16--0.22)} \\ \textcolor{gray}{7.3\,{\tiny (6.4--11.4)}}} & \makecell{0.18\,{\tiny (0.12--0.29)} \\ \textcolor{gray}{10.0\,{\tiny (8.8--10.9)}}} & \makecell{0.27\,{\tiny (0.18--0.30)} \\ \textcolor{gray}{10.5\,{\tiny (9.0--11.6)}}} & \makecell{0.15\,{\tiny (0.13--0.18)} \\ \textcolor{gray}{6.0\,{\tiny (4.9--6.6)}}} & \makecell{0.30\,{\tiny (0.27--0.38)} \\ \textcolor{gray}{11.8\,{\tiny (8.6--16.5)}}} & \makecell{0.18\,{\tiny (0.14--0.26)} \\ \textcolor{gray}{10.8\,{\tiny (9.1--13.9)}}} & \makecell{0.26\,{\tiny (0.15--0.35)} \\ \textcolor{gray}{10.9\,{\tiny (9.7--13.2)}}} & \makecell{0.15\,{\tiny (0.11--0.20)} \\ \textcolor{gray}{9.4\,{\tiny (7.6--12.5)}}} & \makecell{0.20\,{\tiny (0.14--0.28)} \\ \textcolor{gray}{10.4\,{\tiny (8.5--13.2)}}} \\
Chair & \makecell{0.19\,{\tiny (0.18--0.19)}\rlap{$^{\dagger}$} \\ \textcolor{gray}{8.7\,{\tiny (8.3--9.1)}}} & -- & -- & -- & -- & -- & -- & -- & -- & -- & \makecell{0.21\,{\tiny (0.17--0.25)} \\ \textcolor{gray}{12.8\,{\tiny (10.5--15.1)}}} \\
Sitting & \makecell{0.26\,{\tiny (0.22--0.36)} \\ \textcolor{gray}{24.7\,{\tiny (13.6--33.2)}}} & -- & \makecell{0.16\rlap{$^{\dagger}$} \\ \textcolor{gray}{14.9}} & \makecell{0.23\,{\tiny (0.18--0.27)} \\ \textcolor{gray}{20.2\,{\tiny (10.6--26.5)}}} & \makecell{0.12\,{\tiny (0.12--0.12)} \\ \textcolor{gray}{9.4\,{\tiny (9.0--9.9)}}} & \makecell{0.18\,{\tiny (0.17--0.22)} \\ \textcolor{gray}{9.9\,{\tiny (9.1--10.7)}}} & \makecell{0.55\,{\tiny (0.44--0.55)} \\ \textcolor{gray}{18.4\,{\tiny (17.5--18.6)}}} & \makecell{0.18\,{\tiny (0.17--0.18)} \\ \textcolor{gray}{9.4\,{\tiny (9.3--10.4)}}} & \makecell{0.26\,{\tiny (0.20--0.29)} \\ \textcolor{gray}{10.8\,{\tiny (8.6--13.7)}}} & \makecell{0.15\,{\tiny (0.15--0.16)} \\ \textcolor{gray}{9.3\,{\tiny (9.2--9.4)}}} & \makecell{0.23\,{\tiny (0.18--0.32)} \\ \textcolor{gray}{14.9\,{\tiny (10.3--27.1)}}} \\
Footwear & -- & -- & -- & \makecell{0.15\,{\tiny (0.14--0.17)} \\ \textcolor{gray}{5.3\,{\tiny (5.2--5.8)}}} & -- & -- & \makecell{0.23\,{\tiny (0.22--0.28)} \\ \textcolor{gray}{6.4\,{\tiny (6.2--8.2)}}} & \makecell{0.25\,{\tiny (0.25--0.27)} \\ \textcolor{gray}{9.4\,{\tiny (8.9--9.4)}}} & \makecell{0.34\,{\tiny (0.30--0.35)} \\ \textcolor{gray}{10.2\,{\tiny (9.4--10.4)}}} & \makecell{0.35\,{\tiny (0.32--0.37)} \\ \textcolor{gray}{11.6\,{\tiny (10.8--12.0)}}} & \makecell{0.26\,{\tiny (0.20--0.30)} \\ \textcolor{gray}{8.8\,{\tiny (6.6--10.2)}}} \\
Lying & \makecell{0.33\,{\tiny (0.24--0.41)} \\ \textcolor{gray}{22.1\,{\tiny (17.1--26.1)}}} & -- & -- & \makecell{0.21\,{\tiny (0.17--0.27)} \\ \textcolor{gray}{16.5\,{\tiny (13.8--20.8)}}} & \makecell{0.22\,{\tiny (0.15--0.30)} \\ \textcolor{gray}{14.7\,{\tiny (11.5--20.2)}}} & \makecell{0.23\,{\tiny (0.20--0.35)} \\ \textcolor{gray}{16.6\,{\tiny (12.2--23.4)}}} & \makecell{0.30\,{\tiny (0.26--0.36)} \\ \textcolor{gray}{20.3\,{\tiny (17.1--23.1)}}} & \makecell{0.21\,{\tiny (0.17--0.28)} \\ \textcolor{gray}{19.5\,{\tiny (15.2--22.5)}}} & \makecell{0.23\,{\tiny (0.15--0.33)} \\ \textcolor{gray}{19.2\,{\tiny (14.8--23.3)}}} & \makecell{0.92\,{\tiny (0.79--1.04)}\rlap{$^{\dagger}$} \\ \textcolor{gray}{25.3\,{\tiny (23.8--26.7)}}} & \makecell{0.27\,{\tiny (0.19--0.37)} \\ \textcolor{gray}{20.6\,{\tiny (16.1--25.9)}}} \\
Stair & -- & -- & -- & \makecell{0.24\rlap{$^{\dagger}$} \\ \textcolor{gray}{8.4}} & -- & -- & -- & \makecell{0.32\rlap{$^{\dagger}$} \\ \textcolor{gray}{11.9}} & \makecell{0.27\rlap{$^{\dagger}$} \\ \textcolor{gray}{7.6}} & -- & \makecell{0.27\,{\tiny (0.26--0.29)} \\ \textcolor{gray}{8.4\,{\tiny (8.0--10.1)}}} \\
Muscle Stretching & -- & -- & -- & \makecell{0.35\,{\tiny (0.30--0.41)} \\ \textcolor{gray}{11.7\,{\tiny (10.1--17.5)}}} & \makecell{0.35\,{\tiny (0.23--0.36)} \\ \textcolor{gray}{14.2\,{\tiny (12.6--19.6)}}} & \makecell{0.75\rlap{$^{\dagger}$} \\ \textcolor{gray}{17.2}} & -- & \makecell{0.31\,{\tiny (0.29--0.37)} \\ \textcolor{gray}{14.3\,{\tiny (12.8--21.3)}}} & \makecell{0.23\,{\tiny (0.19--0.37)} \\ \textcolor{gray}{11.9\,{\tiny (8.5--14.7)}}} & -- & \makecell{0.29\,{\tiny (0.21--0.36)} \\ \textcolor{gray}{13.5\,{\tiny (10.3--17.3)}}} \\
Sports & -- & -- & -- & \makecell{0.25\,{\tiny (0.20--0.38)} \\ \textcolor{gray}{13.2\,{\tiny (9.3--16.0)}}} & \makecell{0.30\,{\tiny (0.23--0.37)} \\ \textcolor{gray}{15.4\,{\tiny (11.5--18.4)}}} & \makecell{0.33\,{\tiny (0.24--0.43)} \\ \textcolor{gray}{17.7\,{\tiny (10.1--35.6)}}} & \makecell{0.34\,{\tiny (0.28--0.42)} \\ \textcolor{gray}{12.6\,{\tiny (10.2--16.3)}}} & \makecell{0.29\,{\tiny (0.23--0.38)} \\ \textcolor{gray}{15.2\,{\tiny (11.4--20.1)}}} & \makecell{0.31\,{\tiny (0.23--0.38)} \\ \textcolor{gray}{11.8\,{\tiny (9.9--15.2)}}} & \makecell{0.26\,{\tiny (0.19--0.36)} \\ \textcolor{gray}{13.3\,{\tiny (9.3--21.4)}}} & \makecell{0.30\,{\tiny (0.23--0.38)} \\ \textcolor{gray}{13.5\,{\tiny (9.9--17.7)}}} \\
Bed & -- & -- & -- & -- & -- & -- & -- & -- & -- & -- & \makecell{0.30\,{\tiny (0.24--0.35)} \\ \textcolor{gray}{15.2\,{\tiny (14.6--18.6)}}} \\
Putting on Shoes & -- & -- & -- & -- & -- & \makecell{0.31\,{\tiny (0.29--0.32)}\rlap{$^{\dagger}$} \\ \textcolor{gray}{9.8\,{\tiny (9.5--10.1)}}} & -- & -- & -- & -- & \makecell{0.31\,{\tiny (0.29--0.32)}\rlap{$^{\dagger}$} \\ \textcolor{gray}{9.8\,{\tiny (9.5--10.1)}}} \\
Car & -- & -- & -- & -- & -- & -- & -- & -- & -- & -- & \makecell{0.32\,{\tiny (0.30--0.33)} \\ \textcolor{gray}{14.2\,{\tiny (13.6--15.4)}}} \\
Cross-Country Skiing & \makecell{0.33\,{\tiny (0.32--0.36)} \\ \textcolor{gray}{15.0\,{\tiny (13.8--15.6)}}} & -- & -- & -- & -- & -- & -- & -- & -- & -- & \makecell{0.33\,{\tiny (0.32--0.36)} \\ \textcolor{gray}{15.0\,{\tiny (13.8--15.6)}}} \\
Walking & \makecell{0.38\,{\tiny (0.31--0.44)} \\ \textcolor{gray}{13.3\,{\tiny (11.9--14.9)}}} & \makecell{0.42\,{\tiny (0.34--0.48)} \\ \textcolor{gray}{10.7\,{\tiny (9.2--12.4)}}} & \makecell{0.13\,{\tiny (0.11--0.16)} \\ \textcolor{gray}{5.2\,{\tiny (4.7--7.2)}}} & \makecell{0.22\,{\tiny (0.18--0.28)} \\ \textcolor{gray}{8.8\,{\tiny (7.0--11.8)}}} & \makecell{0.18\,{\tiny (0.15--0.21)} \\ \textcolor{gray}{8.1\,{\tiny (8.0--8.5)}}} & \makecell{0.16\,{\tiny (0.13--0.18)} \\ \textcolor{gray}{6.2\,{\tiny (5.6--7.3)}}} & \makecell{0.27\,{\tiny (0.24--0.28)} \\ \textcolor{gray}{8.6\,{\tiny (7.4--10.4)}}} & \makecell{0.23\,{\tiny (0.19--0.25)} \\ \textcolor{gray}{9.1\,{\tiny (8.6--10.5)}}} & \makecell{0.31\,{\tiny (0.28--0.35)} \\ \textcolor{gray}{9.1\,{\tiny (9.0--9.5)}}} & \makecell{0.18\,{\tiny (0.16--0.26)} \\ \textcolor{gray}{6.9\,{\tiny (6.4--9.3)}}} & \makecell{0.34\,{\tiny (0.25--0.41)} \\ \textcolor{gray}{12.5\,{\tiny (9.9--14.9)}}} \\
Standing & \makecell{0.45\,{\tiny (0.36--0.62)} \\ \textcolor{gray}{18.4\,{\tiny (13.3--35.3)}}} & -- & \makecell{0.15\,{\tiny (0.12--0.18)}\rlap{$^{\dagger}$} \\ \textcolor{gray}{10.1\,{\tiny (9.4--10.7)}}} & \makecell{0.34\,{\tiny (0.29--0.44)} \\ \textcolor{gray}{15.3\,{\tiny (13.0--16.4)}}} & \makecell{0.21\,{\tiny (0.18--0.30)} \\ \textcolor{gray}{8.2\,{\tiny (7.3--12.7)}}} & \makecell{0.30\,{\tiny (0.16--0.45)} \\ \textcolor{gray}{9.3\,{\tiny (8.0--13.7)}}} & \makecell{0.55\,{\tiny (0.34--0.56)} \\ \textcolor{gray}{19.2\,{\tiny (13.9--19.8)}}} & \makecell{0.29\,{\tiny (0.22--0.36)} \\ \textcolor{gray}{15.5\,{\tiny (13.7--18.3)}}} & \makecell{0.39\,{\tiny (0.18--0.52)} \\ \textcolor{gray}{14.4\,{\tiny (14.0--17.5)}}} & \makecell{0.24\,{\tiny (0.24--0.26)} \\ \textcolor{gray}{11.3\,{\tiny (8.9--12.3)}}} & \makecell{0.36\,{\tiny (0.25--0.50)} \\ \textcolor{gray}{14.6\,{\tiny (11.4--20.3)}}} \\
Bath Tub & -- & -- & -- & -- & -- & -- & -- & -- & -- & -- & \makecell{0.38\,{\tiny (0.34--0.42)}\rlap{$^{\dagger}$} \\ \textcolor{gray}{14.9\,{\tiny (13.1--16.7)}}} \\
Stairs & \makecell{0.49\,{\tiny (0.47--0.54)} \\ \textcolor{gray}{15.7\,{\tiny (14.7--17.2)}}} & \makecell{0.48\,{\tiny (0.35--0.61)} \\ \textcolor{gray}{13.0\,{\tiny (9.6--16.6)}}} & -- & -- & -- & \makecell{0.36\,{\tiny (0.30--0.43)}\rlap{$^{\dagger}$} \\ \textcolor{gray}{14.5\,{\tiny (12.7--16.4)}}} & -- & -- & \makecell{0.30\rlap{$^{\dagger}$} \\ \textcolor{gray}{10.5}} & -- & \makecell{0.41\,{\tiny (0.35--0.50)} \\ \textcolor{gray}{15.7\,{\tiny (13.4--17.4)}}} \\
Trampoline & \makecell{0.44\,{\tiny (0.39--0.54)} \\ \textcolor{gray}{15.1\,{\tiny (12.4--16.9)}}} & -- & -- & -- & -- & -- & -- & -- & -- & -- & \makecell{0.44\,{\tiny (0.39--0.54)} \\ \textcolor{gray}{15.1\,{\tiny (12.4--16.9)}}} \\
Dance & -- & -- & -- & \makecell{0.55\,{\tiny (0.55--0.55)}\rlap{$^{\dagger}$} \\ \textcolor{gray}{14.0\,{\tiny (13.9--14.1)}}} & -- & -- & -- & -- & -- & -- & \makecell{0.55\,{\tiny (0.55--0.55)}\rlap{$^{\dagger}$} \\ \textcolor{gray}{14.0\,{\tiny (13.9--14.1)}}} \\
Agriculture & -- & -- & -- & \makecell{0.64\,{\tiny (0.52--0.73)} \\ \textcolor{gray}{15.7\,{\tiny (14.0--17.0)}}} & -- & -- & -- & -- & -- & -- & \makecell{0.64\,{\tiny (0.52--0.73)} \\ \textcolor{gray}{15.7\,{\tiny (14.0--17.0)}}} \\
Stumbling & \makecell{0.92\,{\tiny (0.89--0.95)}\rlap{$^{\dagger}$} \\ \textcolor{gray}{13.7\,{\tiny (13.3--14.1)}}} & -- & -- & -- & -- & -- & -- & -- & -- & -- & \makecell{0.66\,{\tiny (0.40--0.95)} \\ \textcolor{gray}{14.3\,{\tiny (13.2--14.7)}}} \\
Muscle Contraction & -- & -- & -- & -- & -- & \makecell{1.16\rlap{$^{\dagger}$} \\ \textcolor{gray}{31.8}} & -- & \makecell{0.81\rlap{$^{\dagger}$} \\ \textcolor{gray}{19.5}} & -- & -- & \makecell{0.98\,{\tiny (0.90--1.07)}\rlap{$^{\dagger}$} \\ \textcolor{gray}{25.6\,{\tiny (22.6--28.7)}}} \\
\midrule
\textbf{All} & \makecell{0.37\,{\tiny (0.28--0.47)} \\ \textcolor{gray}{15.7\,{\tiny (12.9--23.6)}}} & \makecell{0.42\,{\tiny (0.35--0.51)} \\ \textcolor{gray}{10.7\,{\tiny (9.4--12.5)}}} & \makecell{0.18\,{\tiny (0.14--0.21)} \\ \textcolor{gray}{8.2\,{\tiny (6.1--11.2)}}} & \makecell{0.25\,{\tiny (0.17--0.38)} \\ \textcolor{gray}{12.3\,{\tiny (9.3--16.1)}}} & \makecell{0.24\,{\tiny (0.17--0.32)} \\ \textcolor{gray}{12.5\,{\tiny (9.3--16.7)}}} & \makecell{0.23\,{\tiny (0.18--0.36)} \\ \textcolor{gray}{13.0\,{\tiny (8.5--19.5)}}} & \makecell{0.30\,{\tiny (0.23--0.38)} \\ \textcolor{gray}{13.9\,{\tiny (9.4--17.9)}}} & \makecell{0.26\,{\tiny (0.20--0.34)} \\ \textcolor{gray}{13.8\,{\tiny (10.6--19.5)}}} & \makecell{0.29\,{\tiny (0.22--0.36)} \\ \textcolor{gray}{11.9\,{\tiny (9.8--15.3)}}} & \makecell{0.23\,{\tiny (0.16--0.33)} \\ \textcolor{gray}{11.0\,{\tiny (8.9--15.6)}}} & \makecell{0.28\,{\tiny (0.20--0.39)} \\ \textcolor{gray}{14.1\,{\tiny (10.4--18.7)}}} \\
\bottomrule
\end{tabular}

\end{sidewaystable}

\begin{sidewaystable}
  \centering
  \scriptsize
  \setlength{\tabcolsep}{3pt}
  \renewcommand{\arraystretch}{1.1}
  \caption{Hip cohort (part 2/2): continuation of Table~\ref{tab:rmse-hip-1}. Conventions as in part 1. $^\dagger$Based on fewer than 3 trials.}
  \label{tab:rmse-hip-2}
  \begin{tabular}{lcccccccccc}
\toprule
 & \textbf{H9L} & \textbf{H10R} & \textbf{HSR} & \textbf{IBL} & \textbf{JBR} & \textbf{KWL} & \textbf{KWR} & \textbf{PFL} & \textbf{RHR} & \textbf{All} \\
\midrule
Vibration & -- & -- & -- & -- & -- & -- & -- & -- & -- & \makecell{0.18\,{\tiny (0.16--0.23)} \\ \textcolor{gray}{14.6\,{\tiny (11.8--18.1)}}} \\
Bicycle & -- & -- & \makecell{0.17\,{\tiny (0.16--0.18)} \\ \textcolor{gray}{21.2\,{\tiny (18.0--25.0)}}} & \makecell{0.20\,{\tiny (0.19--0.20)} \\ \textcolor{gray}{23.6\,{\tiny (19.3--28.7)}}} & -- & \makecell{0.25\rlap{$^{\dagger}$} \\ \textcolor{gray}{32.7}} & \makecell{0.20\,{\tiny (0.16--0.29)} \\ \textcolor{gray}{21.6\,{\tiny (20.5--24.6)}}} & -- & -- & \makecell{0.19\,{\tiny (0.17--0.22)} \\ \textcolor{gray}{24.1\,{\tiny (19.3--29.2)}}} \\
Gait Analysis & \makecell{0.20\,{\tiny (0.13--0.31)} \\ \textcolor{gray}{13.6\,{\tiny (10.2--15.6)}}} & \makecell{0.19\,{\tiny (0.15--0.23)} \\ \textcolor{gray}{11.4\,{\tiny (9.6--12.5)}}} & -- & -- & -- & -- & -- & -- & -- & \makecell{0.20\,{\tiny (0.14--0.28)} \\ \textcolor{gray}{10.4\,{\tiny (8.5--13.2)}}} \\
Chair & -- & -- & \makecell{0.16\,{\tiny (0.13--0.18)} \\ \textcolor{gray}{10.3\,{\tiny (9.6--13.0)}}} & \makecell{0.25\,{\tiny (0.23--0.31)} \\ \textcolor{gray}{13.1\,{\tiny (11.6--15.4)}}} & \makecell{0.29\,{\tiny (0.20--0.39)} \\ \textcolor{gray}{10.7\,{\tiny (9.4--12.4)}}} & \makecell{0.27\,{\tiny (0.25--0.28)}\rlap{$^{\dagger}$} \\ \textcolor{gray}{13.4\,{\tiny (13.0--13.7)}}} & \makecell{0.23\,{\tiny (0.20--0.26)} \\ \textcolor{gray}{14.8\,{\tiny (14.1--16.7)}}} & \makecell{0.18\,{\tiny (0.16--0.23)} \\ \textcolor{gray}{13.0\,{\tiny (11.7--13.8)}}} & -- & \makecell{0.21\,{\tiny (0.17--0.25)} \\ \textcolor{gray}{12.8\,{\tiny (10.5--15.1)}}} \\
Sitting & \makecell{0.17\,{\tiny (0.16--0.17)} \\ \textcolor{gray}{12.7\,{\tiny (10.9--13.1)}}} & \makecell{0.23\,{\tiny (0.21--0.24)}\rlap{$^{\dagger}$} \\ \textcolor{gray}{13.3\,{\tiny (12.1--14.6)}}} & -- & -- & -- & -- & -- & -- & -- & \makecell{0.23\,{\tiny (0.18--0.32)} \\ \textcolor{gray}{14.9\,{\tiny (10.3--27.1)}}} \\
Footwear & \makecell{0.21\,{\tiny (0.21--0.24)} \\ \textcolor{gray}{7.8\,{\tiny (7.2--8.7)}}} & -- & -- & -- & -- & -- & -- & -- & -- & \makecell{0.26\,{\tiny (0.20--0.30)} \\ \textcolor{gray}{8.8\,{\tiny (6.6--10.2)}}} \\
Lying & -- & -- & \makecell{0.30\,{\tiny (0.22--0.38)} \\ \textcolor{gray}{22.7\,{\tiny (20.3--33.5)}}} & \makecell{0.29\,{\tiny (0.21--0.34)} \\ \textcolor{gray}{21.5\,{\tiny (19.2--24.5)}}} & \makecell{0.23\,{\tiny (0.20--0.26)} \\ \textcolor{gray}{17.7\,{\tiny (14.4--22.4)}}} & \makecell{0.33\,{\tiny (0.31--0.58)} \\ \textcolor{gray}{27.4\,{\tiny (25.5--32.9)}}} & \makecell{0.37\,{\tiny (0.30--0.47)} \\ \textcolor{gray}{22.9\,{\tiny (20.2--25.6)}}} & \makecell{0.20\,{\tiny (0.15--0.26)} \\ \textcolor{gray}{19.2\,{\tiny (16.1--25.5)}}} & -- & \makecell{0.27\,{\tiny (0.19--0.37)} \\ \textcolor{gray}{20.6\,{\tiny (16.1--25.9)}}} \\
Stair & -- & -- & -- & -- & -- & -- & -- & -- & -- & \makecell{0.27\,{\tiny (0.26--0.29)} \\ \textcolor{gray}{8.4\,{\tiny (8.0--10.1)}}} \\
Muscle Stretching & \makecell{0.24\,{\tiny (0.22--0.28)} \\ \textcolor{gray}{12.9\,{\tiny (8.0--16.1)}}} & \makecell{0.24\,{\tiny (0.18--0.29)} \\ \textcolor{gray}{13.3\,{\tiny (10.3--14.2)}}} & -- & -- & -- & -- & -- & -- & -- & \makecell{0.29\,{\tiny (0.21--0.36)} \\ \textcolor{gray}{13.5\,{\tiny (10.3--17.3)}}} \\
Sports & \makecell{0.28\,{\tiny (0.24--0.34)} \\ \textcolor{gray}{10.9\,{\tiny (7.9--14.8)}}} & \makecell{0.27\,{\tiny (0.24--0.40)} \\ \textcolor{gray}{15.5\,{\tiny (13.8--17.5)}}} & -- & -- & -- & -- & -- & -- & -- & \makecell{0.30\,{\tiny (0.23--0.38)} \\ \textcolor{gray}{13.5\,{\tiny (9.9--17.7)}}} \\
Bed & -- & -- & \makecell{0.29\,{\tiny (0.25--0.33)}\rlap{$^{\dagger}$} \\ \textcolor{gray}{19.8\,{\tiny (17.2--22.3)}}} & \makecell{0.33\,{\tiny (0.31--0.34)}\rlap{$^{\dagger}$} \\ \textcolor{gray}{14.8\,{\tiny (14.6--15.0)}}} & -- & -- & \makecell{0.27\,{\tiny (0.26--0.28)}\rlap{$^{\dagger}$} \\ \textcolor{gray}{16.3\,{\tiny (15.7--16.8)}}} & \makecell{0.27\,{\tiny (0.22--0.31)}\rlap{$^{\dagger}$} \\ \textcolor{gray}{18.5\,{\tiny (16.4--20.6)}}} & -- & \makecell{0.30\,{\tiny (0.24--0.35)} \\ \textcolor{gray}{15.2\,{\tiny (14.6--18.6)}}} \\
Putting on Shoes & -- & -- & -- & -- & -- & -- & -- & -- & -- & \makecell{0.31\,{\tiny (0.29--0.32)}\rlap{$^{\dagger}$} \\ \textcolor{gray}{9.8\,{\tiny (9.5--10.1)}}} \\
Car & -- & -- & \makecell{0.31\,{\tiny (0.28--0.33)} \\ \textcolor{gray}{14.7\,{\tiny (13.6--15.8)}}} & -- & -- & -- & \makecell{0.33\,{\tiny (0.32--0.33)} \\ \textcolor{gray}{14.9\,{\tiny (13.9--15.6)}}} & \makecell{0.31\,{\tiny (0.30--0.32)} \\ \textcolor{gray}{13.8\,{\tiny (13.5--14.2)}}} & -- & \makecell{0.32\,{\tiny (0.30--0.33)} \\ \textcolor{gray}{14.2\,{\tiny (13.6--15.4)}}} \\
Cross-Country Skiing & -- & -- & -- & -- & -- & -- & -- & -- & -- & \makecell{0.33\,{\tiny (0.32--0.36)} \\ \textcolor{gray}{15.0\,{\tiny (13.8--15.6)}}} \\
Walking & \makecell{0.17\,{\tiny (0.16--0.19)} \\ \textcolor{gray}{5.8\,{\tiny (5.5--6.4)}}} & \makecell{0.19\,{\tiny (0.16--0.25)} \\ \textcolor{gray}{8.4\,{\tiny (6.9--10.7)}}} & \makecell{0.29\,{\tiny (0.23--0.34)} \\ \textcolor{gray}{11.3\,{\tiny (10.4--15.5)}}} & \makecell{0.36\,{\tiny (0.27--0.40)} \\ \textcolor{gray}{13.9\,{\tiny (12.1--16.2)}}} & \makecell{0.40\,{\tiny (0.37--0.49)} \\ \textcolor{gray}{13.9\,{\tiny (11.3--16.6)}}} & \makecell{0.34\,{\tiny (0.23--0.36)} \\ \textcolor{gray}{14.3\,{\tiny (11.1--15.8)}}} & \makecell{0.29\,{\tiny (0.24--0.36)} \\ \textcolor{gray}{12.5\,{\tiny (10.3--16.9)}}} & \makecell{0.37\,{\tiny (0.29--0.41)} \\ \textcolor{gray}{15.3\,{\tiny (13.6--17.0)}}} & \makecell{0.43\,{\tiny (0.36--0.56)} \\ \textcolor{gray}{15.2\,{\tiny (13.8--19.1)}}} & \makecell{0.34\,{\tiny (0.25--0.41)} \\ \textcolor{gray}{12.5\,{\tiny (9.9--14.9)}}} \\
Standing & \makecell{0.26\,{\tiny (0.21--0.57)} \\ \textcolor{gray}{12.0\,{\tiny (10.4--15.9)}}} & \makecell{0.10\,{\tiny (0.09--0.13)} \\ \textcolor{gray}{6.3\,{\tiny (6.1--6.7)}}} & \makecell{0.31\,{\tiny (0.30--0.34)} \\ \textcolor{gray}{14.8\,{\tiny (13.5--15.6)}}} & \makecell{0.23\rlap{$^{\dagger}$} \\ \textcolor{gray}{9.1}} & \makecell{0.69\,{\tiny (0.42--1.14)} \\ \textcolor{gray}{18.2\,{\tiny (16.4--22.5)}}} & \makecell{0.50\rlap{$^{\dagger}$} \\ \textcolor{gray}{17.7}} & \makecell{0.30\,{\tiny (0.29--0.42)} \\ \textcolor{gray}{12.9\,{\tiny (11.7--14.6)}}} & \makecell{0.42\,{\tiny (0.30--0.48)} \\ \textcolor{gray}{20.1\,{\tiny (14.4--25.9)}}} & -- & \makecell{0.36\,{\tiny (0.25--0.50)} \\ \textcolor{gray}{14.6\,{\tiny (11.4--20.3)}}} \\
Bath Tub & -- & -- & \makecell{0.47\rlap{$^{\dagger}$} \\ \textcolor{gray}{18.4}} & -- & -- & -- & -- & \makecell{0.30\rlap{$^{\dagger}$} \\ \textcolor{gray}{11.3}} & -- & \makecell{0.38\,{\tiny (0.34--0.42)}\rlap{$^{\dagger}$} \\ \textcolor{gray}{14.9\,{\tiny (13.1--16.7)}}} \\
Stairs & -- & -- & \makecell{0.34\,{\tiny (0.27--0.36)} \\ \textcolor{gray}{14.4\,{\tiny (12.2--15.9)}}} & \makecell{0.54\,{\tiny (0.50--0.55)} \\ \textcolor{gray}{15.7\,{\tiny (15.3--16.1)}}} & \makecell{0.80\,{\tiny (0.37--0.89)} \\ \textcolor{gray}{15.7\,{\tiny (12.8--17.2)}}} & \makecell{0.43\,{\tiny (0.42--0.45)}\rlap{$^{\dagger}$} \\ \textcolor{gray}{17.1\,{\tiny (16.8--17.5)}}} & \makecell{0.32\,{\tiny (0.29--0.35)} \\ \textcolor{gray}{13.7\,{\tiny (12.2--17.2)}}} & \makecell{0.39\,{\tiny (0.36--0.42)} \\ \textcolor{gray}{18.5\,{\tiny (14.8--19.3)}}} & \makecell{0.44\,{\tiny (0.40--0.47)} \\ \textcolor{gray}{16.0\,{\tiny (15.5--16.6)}}} & \makecell{0.41\,{\tiny (0.35--0.50)} \\ \textcolor{gray}{15.7\,{\tiny (13.4--17.4)}}} \\
Trampoline & -- & -- & -- & -- & -- & -- & -- & -- & -- & \makecell{0.44\,{\tiny (0.39--0.54)} \\ \textcolor{gray}{15.1\,{\tiny (12.4--16.9)}}} \\
Dance & -- & -- & -- & -- & -- & -- & -- & -- & -- & \makecell{0.55\,{\tiny (0.55--0.55)}\rlap{$^{\dagger}$} \\ \textcolor{gray}{14.0\,{\tiny (13.9--14.1)}}} \\
Agriculture & -- & -- & -- & -- & -- & -- & -- & -- & -- & \makecell{0.64\,{\tiny (0.52--0.73)} \\ \textcolor{gray}{15.7\,{\tiny (14.0--17.0)}}} \\
Stumbling & -- & -- & \makecell{0.38\,{\tiny (0.34--0.42)} \\ \textcolor{gray}{14.0\,{\tiny (11.8--14.4)}}} & -- & \makecell{1.38\rlap{$^{\dagger}$} \\ \textcolor{gray}{17.2}} & -- & -- & -- & -- & \makecell{0.66\,{\tiny (0.40--0.95)} \\ \textcolor{gray}{14.3\,{\tiny (13.2--14.7)}}} \\
Muscle Contraction & -- & -- & -- & -- & -- & -- & -- & -- & -- & \makecell{0.98\,{\tiny (0.90--1.07)}\rlap{$^{\dagger}$} \\ \textcolor{gray}{25.6\,{\tiny (22.6--28.7)}}} \\
\midrule
\textbf{All} & \makecell{0.25\,{\tiny (0.17--0.32)} \\ \textcolor{gray}{11.3\,{\tiny (8.3--15.3)}}} & \makecell{0.24\,{\tiny (0.17--0.28)} \\ \textcolor{gray}{12.7\,{\tiny (9.9--15.5)}}} & \makecell{0.27\,{\tiny (0.19--0.35)} \\ \textcolor{gray}{16.7\,{\tiny (13.2--22.6)}}} & \makecell{0.30\,{\tiny (0.23--0.37)} \\ \textcolor{gray}{16.6\,{\tiny (13.9--21.0)}}} & \makecell{0.40\,{\tiny (0.32--0.62)} \\ \textcolor{gray}{14.9\,{\tiny (12.5--17.4)}}} & \makecell{0.34\,{\tiny (0.26--0.49)} \\ \textcolor{gray}{22.0\,{\tiny (15.9--30.6)}}} & \makecell{0.30\,{\tiny (0.24--0.38)} \\ \textcolor{gray}{16.8\,{\tiny (12.4--22.6)}}} & \makecell{0.27\,{\tiny (0.19--0.38)} \\ \textcolor{gray}{16.4\,{\tiny (13.8--20.6)}}} & \makecell{0.44\,{\tiny (0.36--0.54)} \\ \textcolor{gray}{15.9\,{\tiny (14.0--17.8)}}} & \makecell{0.28\,{\tiny (0.20--0.39)} \\ \textcolor{gray}{14.1\,{\tiny (10.4--18.7)}}} \\
\bottomrule
\end{tabular}

\end{sidewaystable}

\begin{sidewaystable}
  \centering
  \scriptsize
  \setlength{\tabcolsep}{3pt}
  \renewcommand{\arraystretch}{1.1}
  \caption{Knee cohort: per-activity, per-implant prediction error.
    Each cell shows RMSE median (Q1--Q3) in BW (top) with nRMSE median (Q1--Q3)
    in \% on the gray line below. The ``All'' row and column report marginal
    medians over the pooled trial-level distribution across all 9 implants.}
  \label{tab:rmse-knee}
  \begin{tabular}{lcccccccccc}
\toprule
 & \textbf{K1L} & \textbf{K2L} & \textbf{K3R} & \textbf{K4R} & \textbf{K5R} & \textbf{K6L} & \textbf{K7L} & \textbf{K8L} & \textbf{K9L} & \textbf{All} \\
\midrule
Walking & \makecell{0.20\,{\tiny (0.17--0.22)} \\ \textcolor{gray}{7.3\,{\tiny (6.3--8.0)}}} & \makecell{0.18\,{\tiny (0.17--0.23)} \\ \textcolor{gray}{7.5\,{\tiny (6.5--8.7)}}} & \makecell{0.18\,{\tiny (0.16--0.18)} \\ \textcolor{gray}{6.7\,{\tiny (6.0--7.1)}}} & \makecell{0.22\,{\tiny (0.21--0.22)}\rlap{$^{\dagger}$} \\ \textcolor{gray}{7.2\,{\tiny (7.1--7.3)}}} & \makecell{0.18\,{\tiny (0.17--0.20)} \\ \textcolor{gray}{7.5\,{\tiny (7.0--8.2)}}} & -- & \makecell{0.15\,{\tiny (0.12--0.16)} \\ \textcolor{gray}{5.3\,{\tiny (4.4--5.6)}}} & \makecell{0.14\,{\tiny (0.12--0.15)} \\ \textcolor{gray}{5.7\,{\tiny (4.9--6.2)}}} & \makecell{0.15\,{\tiny (0.15--0.16)} \\ \textcolor{gray}{7.2\,{\tiny (6.9--7.6)}}} & \makecell{0.17\,{\tiny (0.15--0.20)} \\ \textcolor{gray}{6.9\,{\tiny (6.1--7.9)}}} \\
Vibration & \makecell{0.18\,{\tiny (0.14--0.25)} \\ \textcolor{gray}{8.7\,{\tiny (7.6--9.0)}}} & -- & \makecell{0.17\,{\tiny (0.15--0.18)} \\ \textcolor{gray}{9.1\,{\tiny (7.5--10.0)}}} & -- & \makecell{0.17\,{\tiny (0.15--0.22)} \\ \textcolor{gray}{8.4\,{\tiny (7.7--9.2)}}} & -- & \makecell{0.18\,{\tiny (0.11--0.24)} \\ \textcolor{gray}{8.8\,{\tiny (7.3--11.7)}}} & \makecell{0.19\,{\tiny (0.16--0.20)} \\ \textcolor{gray}{9.8\,{\tiny (8.3--10.6)}}} & \makecell{0.16\,{\tiny (0.14--0.19)} \\ \textcolor{gray}{7.8\,{\tiny (7.7--10.3)}}} & \makecell{0.17\,{\tiny (0.14--0.22)} \\ \textcolor{gray}{8.7\,{\tiny (7.6--10.4)}}} \\
Standing & \makecell{0.26\,{\tiny (0.19--0.29)} \\ \textcolor{gray}{9.7\,{\tiny (9.0--10.0)}}} & \makecell{0.17\,{\tiny (0.16--0.18)} \\ \textcolor{gray}{7.7\,{\tiny (7.4--8.8)}}} & \makecell{0.19\,{\tiny (0.17--0.19)} \\ \textcolor{gray}{7.4\,{\tiny (7.2--8.4)}}} & \makecell{0.17\rlap{$^{\dagger}$} \\ \textcolor{gray}{13.3}} & \makecell{0.16\,{\tiny (0.16--0.19)} \\ \textcolor{gray}{6.6\,{\tiny (6.0--8.3)}}} & -- & \makecell{0.14\,{\tiny (0.10--0.19)}\rlap{$^{\dagger}$} \\ \textcolor{gray}{6.6\,{\tiny (6.1--7.0)}}} & \makecell{0.20\,{\tiny (0.16--0.24)}\rlap{$^{\dagger}$} \\ \textcolor{gray}{10.8\,{\tiny (9.8--11.7)}}} & \makecell{0.08\rlap{$^{\dagger}$} \\ \textcolor{gray}{8.3}} & \makecell{0.18\,{\tiny (0.14--0.21)} \\ \textcolor{gray}{7.8\,{\tiny (7.1--10.2)}}} \\
Sports & \makecell{0.10\,{\tiny (0.10--0.12)} \\ \textcolor{gray}{6.9\,{\tiny (6.8--7.3)}}} & \makecell{0.12\,{\tiny (0.11--0.16)} \\ \textcolor{gray}{11.3\,{\tiny (9.1--15.2)}}} & \makecell{0.18\,{\tiny (0.16--0.20)} \\ \textcolor{gray}{12.0\,{\tiny (10.4--12.8)}}} & \makecell{0.13\,{\tiny (0.12--0.22)} \\ \textcolor{gray}{19.0\,{\tiny (15.3--21.9)}}} & \makecell{0.21\,{\tiny (0.16--0.26)} \\ \textcolor{gray}{13.5\,{\tiny (11.8--23.2)}}} & \makecell{0.14\,{\tiny (0.13--0.17)} \\ \textcolor{gray}{9.0\,{\tiny (8.9--10.2)}}} & \makecell{0.19\,{\tiny (0.14--0.26)} \\ \textcolor{gray}{13.8\,{\tiny (9.1--16.3)}}} & \makecell{0.18\,{\tiny (0.14--0.26)} \\ \textcolor{gray}{11.1\,{\tiny (9.5--15.6)}}} & \makecell{0.10\,{\tiny (0.08--0.10)} \\ \textcolor{gray}{7.3\,{\tiny (6.7--8.6)}}} & \makecell{0.18\,{\tiny (0.13--0.24)} \\ \textcolor{gray}{12.5\,{\tiny (10.0--16.5)}}} \\
Sitting & \makecell{0.24\,{\tiny (0.23--0.28)} \\ \textcolor{gray}{9.7\,{\tiny (9.3--10.5)}}} & \makecell{0.16\,{\tiny (0.16--0.17)} \\ \textcolor{gray}{6.4\,{\tiny (6.2--6.9)}}} & \makecell{0.17\,{\tiny (0.16--0.18)} \\ \textcolor{gray}{6.5\,{\tiny (6.5--7.2)}}} & \makecell{0.30\,{\tiny (0.30--0.32)} \\ \textcolor{gray}{15.8\,{\tiny (15.5--16.0)}}} & \makecell{0.26\,{\tiny (0.25--0.28)} \\ \textcolor{gray}{10.4\,{\tiny (9.9--10.9)}}} & -- & -- & -- & -- & \makecell{0.24\,{\tiny (0.17--0.30)} \\ \textcolor{gray}{9.5\,{\tiny (7.0--11.3)}}} \\
Stair & -- & -- & \makecell{0.19\rlap{$^{\dagger}$} \\ \textcolor{gray}{6.0}} & -- & \makecell{0.25\rlap{$^{\dagger}$} \\ \textcolor{gray}{7.6}} & -- & -- & \makecell{0.36\rlap{$^{\dagger}$} \\ \textcolor{gray}{11.7}} & -- & \makecell{0.25\,{\tiny (0.22--0.31)} \\ \textcolor{gray}{7.6\,{\tiny (6.8--9.7)}}} \\
Gait Analysis & \makecell{0.32\,{\tiny (0.26--0.38)} \\ \textcolor{gray}{9.3\,{\tiny (8.7--10.8)}}} & \makecell{0.19\,{\tiny (0.13--0.22)} \\ \textcolor{gray}{6.5\,{\tiny (4.9--7.4)}}} & \makecell{0.27\,{\tiny (0.18--0.33)} \\ \textcolor{gray}{10.0\,{\tiny (6.9--14.3)}}} & \makecell{0.25\,{\tiny (0.23--0.26)} \\ \textcolor{gray}{7.5\,{\tiny (7.2--7.7)}}} & \makecell{0.28\,{\tiny (0.23--0.34)} \\ \textcolor{gray}{8.6\,{\tiny (7.8--9.9)}}} & \makecell{0.28\,{\tiny (0.25--0.31)} \\ \textcolor{gray}{7.6\,{\tiny (7.0--8.0)}}} & \makecell{0.25\,{\tiny (0.21--0.27)} \\ \textcolor{gray}{7.4\,{\tiny (6.3--9.0)}}} & \makecell{0.20\,{\tiny (0.15--0.22)} \\ \textcolor{gray}{6.9\,{\tiny (5.6--7.3)}}} & \makecell{0.27\,{\tiny (0.24--0.33)} \\ \textcolor{gray}{13.3\,{\tiny (12.8--15.7)}}} & \makecell{0.25\,{\tiny (0.20--0.30)} \\ \textcolor{gray}{7.9\,{\tiny (6.8--10.5)}}} \\
Deep Knee Bend & \makecell{0.37\,{\tiny (0.31--0.41)} \\ \textcolor{gray}{12.6\,{\tiny (10.8--13.6)}}} & \makecell{0.17\,{\tiny (0.16--0.17)} \\ \textcolor{gray}{6.6\,{\tiny (6.3--7.2)}}} & \makecell{0.21\,{\tiny (0.21--0.22)} \\ \textcolor{gray}{8.4\,{\tiny (8.3--8.8)}}} & \makecell{0.40\,{\tiny (0.38--0.41)} \\ \textcolor{gray}{23.0\,{\tiny (21.5--23.6)}}} & \makecell{0.32\,{\tiny (0.30--0.34)} \\ \textcolor{gray}{12.3\,{\tiny (12.0--12.5)}}} & -- & \makecell{0.17\rlap{$^{\dagger}$} \\ \textcolor{gray}{6.6}} & \makecell{0.24\rlap{$^{\dagger}$} \\ \textcolor{gray}{9.9}} & -- & \makecell{0.27\,{\tiny (0.21--0.35)} \\ \textcolor{gray}{10.6\,{\tiny (8.2--12.9)}}} \\
Stairs & \makecell{0.31\,{\tiny (0.24--0.34)} \\ \textcolor{gray}{7.9\,{\tiny (7.0--9.7)}}} & \makecell{0.24\,{\tiny (0.21--0.36)} \\ \textcolor{gray}{8.6\,{\tiny (6.7--11.1)}}} & \makecell{0.26\,{\tiny (0.25--0.32)} \\ \textcolor{gray}{8.5\,{\tiny (7.7--9.8)}}} & \makecell{0.33\,{\tiny (0.30--0.35)} \\ \textcolor{gray}{11.0\,{\tiny (9.4--11.1)}}} & \makecell{0.30\,{\tiny (0.27--0.35)} \\ \textcolor{gray}{8.7\,{\tiny (8.0--9.5)}}} & -- & -- & -- & -- & \makecell{0.30\,{\tiny (0.24--0.34)} \\ \textcolor{gray}{8.6\,{\tiny (7.4--10.0)}}} \\
Knee Brace & \makecell{0.36\,{\tiny (0.20--0.41)} \\ \textcolor{gray}{8.9\,{\tiny (7.7--10.8)}}} & -- & \makecell{0.28\,{\tiny (0.20--0.30)} \\ \textcolor{gray}{10.0\,{\tiny (9.0--11.3)}}} & -- & \makecell{0.30\,{\tiny (0.20--0.34)} \\ \textcolor{gray}{9.5\,{\tiny (9.0--9.9)}}} & -- & -- & -- & -- & \makecell{0.30\,{\tiny (0.20--0.34)} \\ \textcolor{gray}{9.5\,{\tiny (8.7--10.9)}}} \\
\midrule
\textbf{All} & \makecell{0.24\,{\tiny (0.19--0.34)} \\ \textcolor{gray}{8.5\,{\tiny (7.4--9.9)}}} & \makecell{0.18\,{\tiny (0.15--0.23)} \\ \textcolor{gray}{7.4\,{\tiny (6.4--8.9)}}} & \makecell{0.19\,{\tiny (0.17--0.26)} \\ \textcolor{gray}{8.6\,{\tiny (7.1--10.6)}}} & \makecell{0.25\,{\tiny (0.18--0.31)} \\ \textcolor{gray}{14.3\,{\tiny (7.9--20.1)}}} & \makecell{0.22\,{\tiny (0.17--0.30)} \\ \textcolor{gray}{9.7\,{\tiny (8.1--12.1)}}} & \makecell{0.28\,{\tiny (0.22--0.31)} \\ \textcolor{gray}{7.8\,{\tiny (7.3--8.6)}}} & \makecell{0.20\,{\tiny (0.15--0.26)} \\ \textcolor{gray}{8.1\,{\tiny (6.5--10.8)}}} & \makecell{0.18\,{\tiny (0.14--0.22)} \\ \textcolor{gray}{8.3\,{\tiny (6.1--11.0)}}} & \makecell{0.18\,{\tiny (0.15--0.27)} \\ \textcolor{gray}{9.3\,{\tiny (7.6--13.1)}}} & \makecell{0.21\,{\tiny (0.16--0.28)} \\ \textcolor{gray}{8.7\,{\tiny (7.2--11.4)}}} \\
\bottomrule
\end{tabular}

\end{sidewaystable}

\clearpage
\begin{table}[t]
    \centering
    \scriptsize
    \caption{Per-trial resultant force metrics on the Grand Challenge dataset. RMSE in units of body weight (BW); $r^2$ is the squared Pearson correlation. Competitions 1 and 4 measure only axial force, so the resultant reduces to $|F_z|$.}
    \label{tab:per-trial-resultant}
    \begin{tabular}{lrr|lrr|lrr}
        \toprule
        \textbf{Trial}       & \textbf{RMSE} & \textbf{$\boldsymbol{r}^2$} & \textbf{Trial}                & \textbf{RMSE} & \textbf{$\boldsymbol{r}^2$} & \textbf{Trial}            & \textbf{RMSE} & \textbf{$\boldsymbol{r}^2$} \\
        \midrule
        1\_jw\_mtgait\_2     & 0.43          & 0.74                        & 3\_sc\_ngait\_og9             & 0.55          & 0.80                        & 4\_jw\_wpgait\_lw1        & 0.41          & 0.84                        \\
        1\_jw\_mtgait\_10    & 0.36          & 0.80                        & 3\_sc\_smooth\_og1            & 0.44          & 0.87                        & 4\_jw\_wpgait\_lw4        & 0.49          & 0.75                        \\
        1\_jw\_mtgait\_12    & 0.50          & 0.66                        & 3\_sc\_smooth\_og2            & 0.43          & 0.90                        & 4\_jw\_wpgait\_lw5        & 0.50          & 0.74                        \\
        1\_jw\_mtgait\_13    & 0.37          & 0.81                        & 3\_sc\_smooth\_og3            & 0.43          & 0.86                        & 4\_jw\_wpgait\_lw6        & 0.51          & 0.72                        \\
        1\_jw\_mtgait\_17    & 0.40          & 0.76                        & 3\_sc\_smooth\_og4            & 0.43          & 0.86                        & 4\_jw\_wpgait\_lw7        & 0.50          & 0.69                        \\
        1\_jw\_ngait\_2      & 0.35          & 0.84                        & 3\_sc\_smooth\_og5            & 0.32          & 0.93                        & 4\_jw\_wpgait\_lw8        & 0.52          & 0.72                        \\
        1\_jw\_ngait\_3      & 0.33          & 0.87                        & 3\_sc\_trunksway1             & 0.34          & 0.92                        & 4\_jw\_wpgait\_sn1        & 0.40          & 0.85                        \\
        1\_jw\_ngait\_4      & 0.35          & 0.84                        & 3\_sc\_trunksway4             & 0.40          & 0.89                        & 4\_jw\_wpgait\_sn3        & 0.43          & 0.80                        \\
        1\_jw\_ngait\_5      & 0.40          & 0.79                        & 3\_sc\_trunksway5             & 0.48          & 0.86                        & 4\_jw\_wpgait\_sn4        & 0.39          & 0.81                        \\
        1\_jw\_ngait\_6      & 0.31          & 0.89                        & 3\_sc\_trunksway6             & 0.54          & 0.88                        & 4\_jw\_wpgait\_sn6        & 0.42          & 0.80                        \\
        1\_jw\_tsgait\_2     & 0.46          & 0.71                        & 3\_sc\_trunksway7             & 0.56          & 0.85                        & 4\_jw\_wpgait\_sn7        & 0.45          & 0.76                        \\
        1\_jw\_tsgait\_3     & 0.42          & 0.76                        & 3\_sc\_wpgait\_l5             & 0.45          & 0.85                        & 4\_jw\_wpgait\_sn8        & 0.41          & 0.82                        \\
        1\_jw\_tsgait\_5     & 0.39          & 0.74                        & 3\_sc\_wpgait\_l8             & 0.53          & 0.75                        & 4\_jw\_wpgait\_sn10       & 0.42          & 0.83                        \\
        1\_jw\_tsgait\_10    & 0.51          & 0.63                        & 3\_sc\_wpgait\_l11            & 0.48          & 0.86                        & 4\_jw\_wpgait\_sw1        & 0.45          & 0.76                        \\
        1\_jw\_tsgait\_11    & 0.49          & 0.68                        & 3\_sc\_wpgait\_s1             & 0.51          & 0.73                        & 4\_jw\_wpgait\_sw2        & 0.45          & 0.86                        \\
        1\_jw\_wpgait\_6     & 0.52          & 0.67                        & 3\_sc\_wpgait\_s5             & 0.43          & 0.87                        & 4\_jw\_wpgait\_sw7        & 0.48          & 0.76                        \\
        1\_jw\_wpgait\_8     & 0.58          & 0.62                        & 3\_sc\_wpgait\_s6             & 0.49          & 0.76                        & 4\_jw\_wpgait\_sw8        & 0.44          & 0.77                        \\
        1\_jw\_wpgait\_10    & 0.55          & 0.48                        & 4\_jw\_bouncy1                & 0.58          & 0.54                        & 4\_jw\_wpgait\_sw9        & 0.35          & 0.88                        \\
        1\_jw\_wpgait\_11    & 0.53          & 0.56                        & 4\_jw\_bouncy4                & 0.34          & 0.86                        & 4\_jw\_wpgait\_sw10       & 0.45          & 0.78                        \\
        1\_jw\_wpgait\_12    & 0.72          & 0.34                        & 4\_jw\_bouncy5                & 0.41          & 0.76                        & 4\_jw\_wpgait\_sw12       & 0.41          & 0.79                        \\
        2\_dm\_mtgait\_3     & 0.67          & 0.69                        & 4\_jw\_bouncy7                & 0.37          & 0.81                        & 5\_ps\_ngait\_og\_ss1     & 0.46          & 0.86                        \\
        2\_dm\_mtgait\_4     & 0.60          & 0.75                        & 4\_jw\_bouncy8                & 0.37          & 0.83                        & 5\_ps\_ngait\_og\_ss3     & 0.56          & 0.78                        \\
        2\_dm\_mtgait\_5     & 0.69          & 0.65                        & 4\_jw\_bouncy9                & 0.36          & 0.86                        & 5\_ps\_ngait\_og\_ss7     & 0.47          & 0.85                        \\
        2\_dm\_mtgait\_6     & 0.66          & 0.70                        & 4\_jw\_medthrust2             & 0.37          & 0.82                        & 5\_ps\_ngait\_og\_ss8     & 0.45          & 0.87                        \\
        2\_dm\_mtgait\_10    & 0.65          & 0.73                        & 4\_jw\_medthrust3             & 0.35          & 0.83                        & 5\_ps\_ngait\_og\_ss9     & 0.40          & 0.88                        \\
        2\_dm\_ngait\_4      & 0.73          & 0.39                        & 4\_jw\_medthrust4             & 0.39          & 0.77                        & 5\_ps\_ngait\_og\_ss11    & 0.44          & 0.87                        \\
        2\_dm\_ngait\_10     & 0.51          & 0.68                        & 4\_jw\_medthrust6             & 0.41          & 0.80                        & 5\_ps\_ngait\_tmf\_ss1hs  & 0.79          & 0.73                        \\
        2\_dm\_ngait\_11     & 0.46          & 0.74                        & 4\_jw\_medthrust11            & 0.38          & 0.80                        & 5\_ps\_rightturn4         & 0.33          & 0.86                        \\
        2\_dm\_ngait\_12     & 0.45          & 0.73                        & 4\_jw\_medthrust12            & 0.40          & 0.81                        & 5\_ps\_rightturn5         & 0.29          & 0.91                        \\
        2\_dm\_ngait\_13     & 0.47          & 0.71                        & 4\_jw\_medthrust13            & 0.46          & 0.74                        & 5\_ps\_rightturn6         & 0.28          & 0.90                        \\
        2\_dm\_tsgait\_1     & 0.58          & 0.60                        & 4\_jw\_medthrust14            & 0.40          & 0.78                        & 6\_dm\_bouncy1            & 0.50          & 0.75                        \\
        2\_dm\_tsgait\_2     & 0.60          & 0.62                        & 4\_jw\_mildcrouch1            & 0.41          & 0.80                        & 6\_dm\_bouncy2            & 0.40          & 0.81                        \\
        2\_dm\_tsgait\_6     & 0.65          & 0.62                        & 4\_jw\_mildcrouch2            & 0.42          & 0.81                        & 6\_dm\_bouncy3            & 0.80          & 0.45                        \\
        2\_dm\_tsgait\_7     & 0.54          & 0.71                        & 4\_jw\_mildcrouch3            & 0.39          & 0.84                        & 6\_dm\_bouncy4            & 0.36          & 0.83                        \\
        2\_dm\_tsgait\_8     & 0.65          & 0.70                        & 4\_jw\_mildcrouch4            & 0.39          & 0.80                        & 6\_dm\_bouncy5            & 0.40          & 0.80                        \\
        2\_dm\_wpgait\_9     & 0.50          & 0.68                        & 4\_jw\_mildcrouch5            & 0.33          & 0.87                        & 6\_dm\_bouncy6            & 0.45          & 0.83                        \\
        2\_dm\_wpgait\_11    & 0.46          & 0.78                        & 4\_jw\_mildcrouch6            & 0.42          & 0.80                        & 6\_dm\_crouch\_og1        & 0.40          & 0.88                        \\
        2\_dm\_wpgait\_12    & 0.40          & 0.80                        & 4\_jw\_moderatecrouch2        & 0.33          & 0.83                        & 6\_dm\_crouch\_og2        & 0.46          & 0.81                        \\
        2\_dm\_wpgait\_13    & 0.50          & 0.72                        & 4\_jw\_moderatecrouch3        & 0.33          & 0.84                        & 6\_dm\_crouch\_og3        & 0.48          & 0.85                        \\
        2\_dm\_wpgait\_17    & 0.45          & 0.73                        & 4\_jw\_moderatecrouch4        & 0.37          & 0.81                        & 6\_dm\_crouch\_og4        & 0.52          & 0.81                        \\
        3\_sc\_bouncy\_og3   & 0.81          & 0.69                        & 4\_jw\_moderatecrouch5        & 0.42          & 0.79                        & 6\_dm\_crouch\_og5        & 0.44          & 0.85                        \\
        3\_sc\_bouncy\_og5   & 1.06          & 0.66                        & 4\_jw\_moderatecrouch6        & 0.34          & 0.85                        & 6\_dm\_crouch\_tm1        & 0.38          & 0.87                        \\
        3\_sc\_bouncy\_og6   & 0.90          & 0.65                        & 4\_jw\_mtpgait2               & 0.35          & 0.84                        & 6\_dm\_mtpgait2           & 0.54          & 0.87                        \\
        3\_sc\_bouncy\_og7   & 0.92          & 0.63                        & 4\_jw\_mtpgait3               & 0.37          & 0.83                        & 6\_dm\_mtpgait3           & 0.35          & 0.91                        \\
        3\_sc\_bouncy\_og8   & 0.88          & 0.61                        & 4\_jw\_mtpgait4               & 0.38          & 0.79                        & 6\_dm\_mtpgait4           & 0.38          & 0.86                        \\
        3\_sc\_crouch\_og1   & 1.07          & 0.81                        & 4\_jw\_mtpgait6               & 0.43          & 0.76                        & 6\_dm\_mtpgait5           & 0.38          & 0.92                        \\
        3\_sc\_crouch\_og3   & 1.10          & 0.66                        & 4\_jw\_mtpgait8               & 0.42          & 0.77                        & 6\_dm\_mtpgait6           & 0.45          & 0.83                        \\
        3\_sc\_crouch\_og4   & 0.98          & 0.75                        & 4\_jw\_mtpgait9               & 0.36          & 0.84                        & 6\_dm\_ngait\_og1         & 0.38          & 0.86                        \\
        3\_sc\_crouch\_og5   & 0.86          & 0.83                        & 4\_jw\_ngait\_og1             & 0.34          & 0.85                        & 6\_dm\_ngait\_og2         & 0.33          & 0.87                        \\
        3\_sc\_crouch\_og6   & 0.84          & 0.71                        & 4\_jw\_ngait\_og2             & 0.39          & 0.78                        & 6\_dm\_ngait\_og3         & 0.28          & 0.89                        \\
        3\_sc\_medialthrust3 & 0.55          & 0.83                        & 4\_jw\_ngait\_og3             & 0.33          & 0.85                        & 6\_dm\_ngait\_og4         & 0.34          & 0.89                        \\
        3\_sc\_medialthrust4 & 0.66          & 0.82                        & 4\_jw\_ngait\_og4             & 0.36          & 0.82                        & 6\_dm\_ngait\_og5         & 0.32          & 0.87                        \\
        3\_sc\_medialthrust5 & 0.59          & 0.80                        & 4\_jw\_ngait\_og5             & 0.50          & 0.64                        & 6\_dm\_ngait\_og6         & 0.38          & 0.84                        \\
        3\_sc\_medialthrust6 & 0.69          & 0.75                        & 4\_jw\_ngait\_og7             & 0.38          & 0.81                        & 6\_dm\_ngait\_og7         & 0.39          & 0.86                        \\
        3\_sc\_medialthrust8 & 0.65          & 0.77                        & 4\_jw\_ngait\_tm\_fast1       & 0.72          & 0.58                        & 6\_dm\_ngait\_og9         & 0.42          & 0.81                        \\
        3\_sc\_mtpgait1      & 0.33          & 0.87                        & 4\_jw\_ngait\_tm\_set1        & 0.79          & 0.61                        & 6\_dm\_ngait\_tm\_med1    & 0.64          & 0.75                        \\
        3\_sc\_mtpgait2      & 0.44          & 0.89                        & 4\_jw\_ngait\_tm\_slow1       & 0.78          & 0.69                        & 6\_dm\_ngait\_tm\_set1    & 0.50          & 0.83                        \\
        3\_sc\_mtpgait3      & 0.44          & 0.86                        & 4\_jw\_ngait\_tm\_ss1         & 0.76          & 0.66                        & 6\_dm\_ngait\_tm\_slow1   & 0.60          & 0.83                        \\
        3\_sc\_mtpgait4      & 0.35          & 0.88                        & 4\_jw\_ngait\_tm\_transition1 & 0.61          & 0.63                        & 6\_dm\_ngait\_tm\_ss1     & 0.47          & 0.85                        \\
        3\_sc\_mtpgait5      & 0.60          & 0.67                        & 4\_jw\_wpgait\_ln2            & 0.39          & 0.85                        & 6\_dm\_ngait\_tmf\_slow1  & 0.53          & 0.84                        \\
        3\_sc\_mtpgait6      & 0.41          & 0.83                        & 4\_jw\_wpgait\_ln4            & 0.42          & 0.82                        & 6\_dm\_ngait\_tmf\_slow2  & 0.54          & 0.82                        \\
        3\_sc\_ngait\_og5    & 0.56          & 0.83                        & 4\_jw\_wpgait\_ln5            & 0.49          & 0.77                        & 6\_dm\_ngait\_transition1 & 0.47          & 0.76                        \\
        3\_sc\_ngait\_og6    & 0.47          & 0.86                        & 4\_jw\_wpgait\_ln6            & 0.51          & 0.77                        & 6\_dm\_smooth1            & 0.47          & 0.72                        \\
        3\_sc\_ngait\_og7    & 0.48          & 0.87                        & 4\_jw\_wpgait\_ln7            & 0.41          & 0.83                        & 6\_dm\_smooth3            & 0.40          & 0.82                        \\
        3\_sc\_ngait\_og8    & 0.60          & 0.70                        & 4\_jw\_wpgait\_ln8            & 0.39          & 0.81                        & 6\_dm\_smooth4            & 0.41          & 0.78                        \\
        \bottomrule
    \end{tabular}
\end{table}

\begin{figure}[p]
  \centering
  \includegraphics[width=\linewidth]{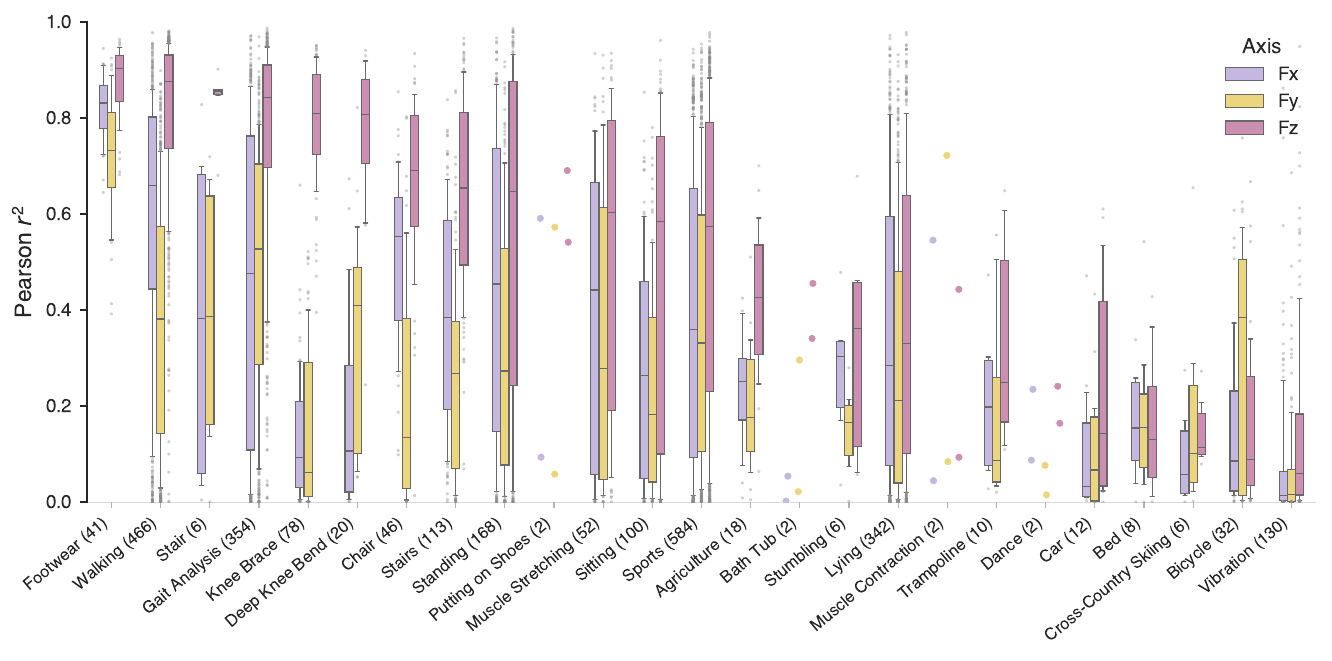}
  \caption{Per-activity temporal shape agreement of force predictions.
    Squared Pearson correlation $r^2$ between predicted and measured
    force traces across all LOSO held-out trials, separately for each
    force component ($F_x$, $F_y$, $F_z$; color-coded). Bounded in
    $[0, 1]$ and invariant to additive offset and multiplicative scale,
    $r^2$ measures how well the predicted trace's temporal shape
    tracks the measured trace, complementing the magnitude error
    reported by per-trial RMSE (Fig.~\ref{fig:raincloud-rmse}).
    Activities are sorted by median $F_z$ $r^2$ (descending). Cyclic,
    weight-bearing tasks (Walking, Stairs, Footwear) achieve the
    highest shape agreement; near-stationary activities (Vibration,
    Lying, Bicycle) yield lower $r^2$ because their target signals
    carry little coherent temporal structure to correlate against,
    not because absolute prediction error is large (compare RMSE in
    Fig.~\ref{fig:raincloud-rmse}). Boxes show median and IQR;
    whiskers span the 10th--90th percentile, outliers shown as small
    dots. Categories with fewer than five trials are rendered as
    individual points.
  }
  \label{fig:r2_box}
\end{figure}

\begin{figure}[p]
  \centering
  \includegraphics[width=\linewidth]{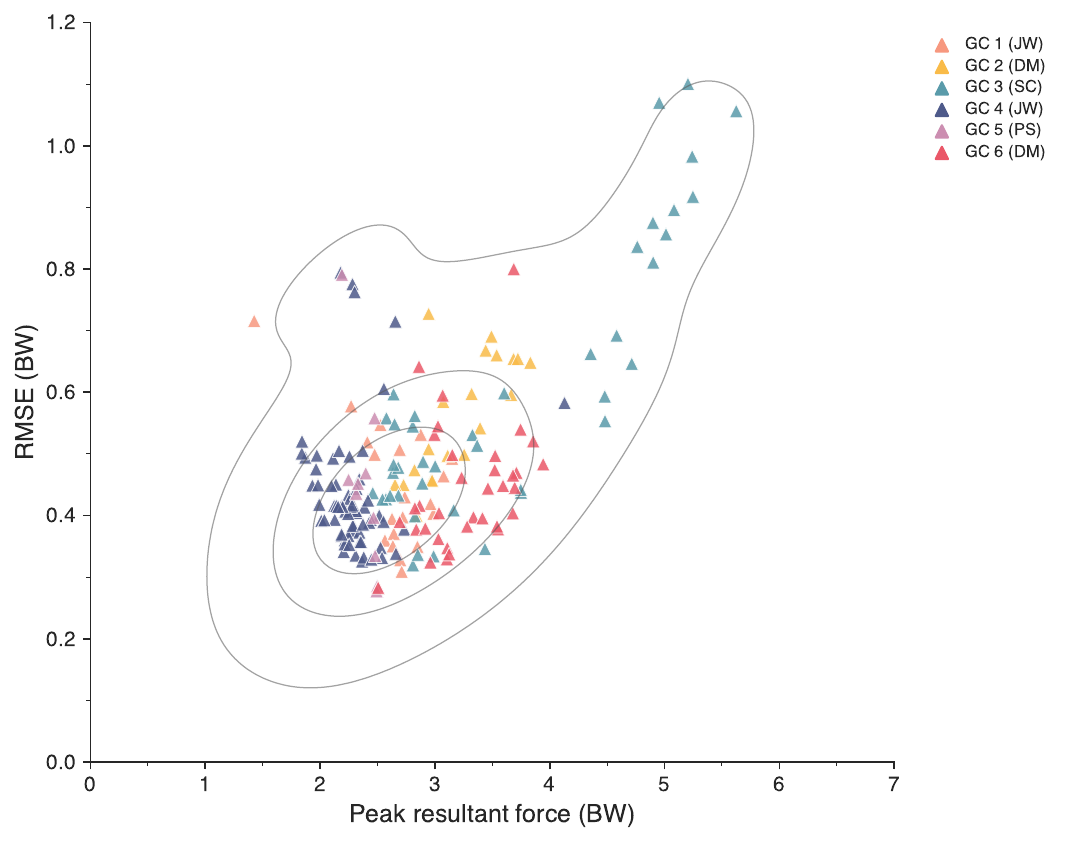}
  \caption{Per-trial prediction error (RMSE) versus peak joint resultant
    force, both expressed in body weights (BW), evaluated on all 195 trials
    of the six Grand Challenge knee implant datasets (four unique patients;
    JW and DM each contributed to two competitions). Each point represents
    one trial, colored by Grand Challenge competition. Contour lines show
    the kernel density estimate of the full dataset. For competitions 1 and
    4 the implant measures only the axial component, so the resultant
    reduces to $|F_z|$.}
  \label{fig:error-vs-peak-force-fregly}
\end{figure}

\begin{figure}[p]
  \centering
  \includegraphics[width=\linewidth]{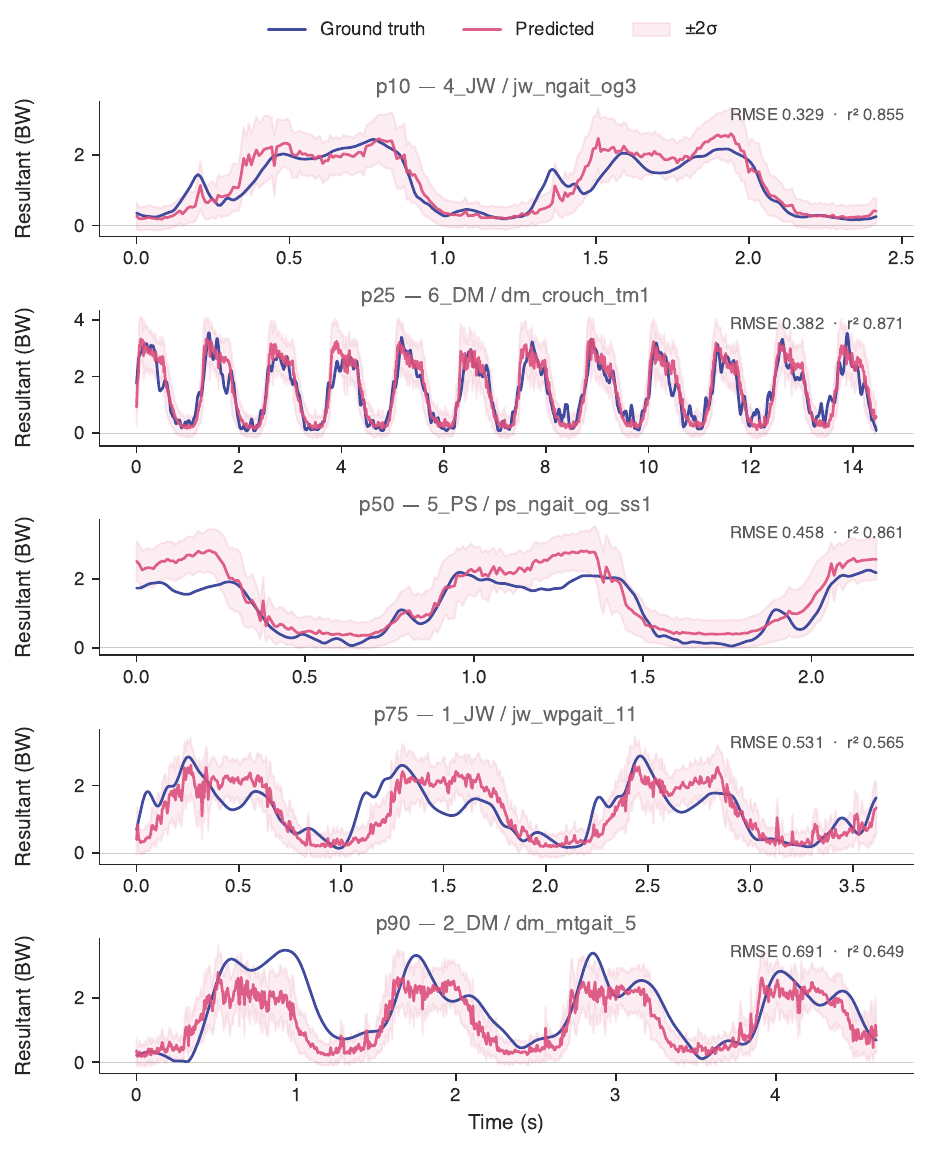}
  \caption{Representative predicted (rose) versus ground-truth (indigo)
    knee implant resultant force traces on the Grand Challenge dataset,
    both expressed in body weights (BW). The five trials are sampled at
    the 10th, 25th, 50th, 75th, and 90th percentiles of the per-trial RMSE
    distribution ($n = 195$), spanning typical-easy (p10) to near-worst (p90)
    cases. Shaded bands give the predicted $\pm 2\sigma$ uncertainty,
    propagated from the per-component variance to the resultant via the
    delta method ($\sigma_R^2 \approx \sum_i (F_i/R)^2 \sigma_i^2$). Each
    panel is annotated with the source competition, trial identifier,
    and per-trial RMSE and $r^2$.}
  \label{fig:gc-percentile-traces}
\end{figure}

\begin{figure}[p]
  \centering
  \includegraphics[width=\linewidth]{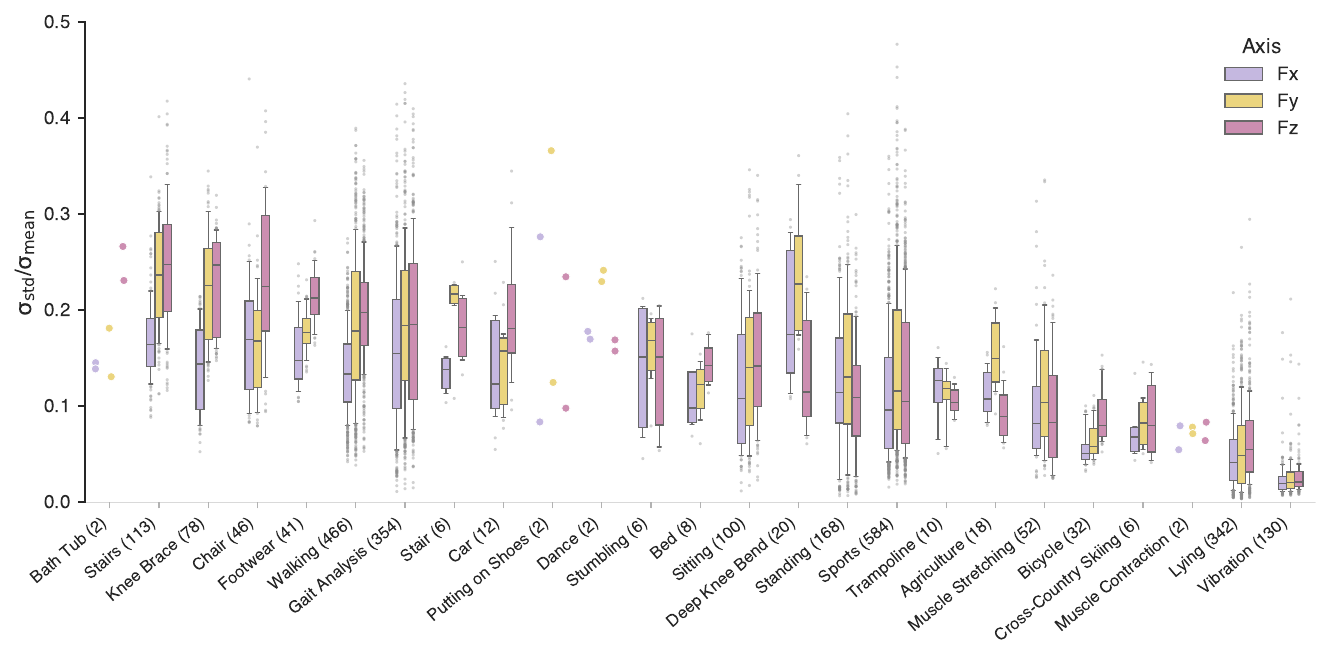}
  \caption{Per-activity heteroscedasticity of predicted uncertainty.
    Within-trial coefficient of variation $\sigma_{\mathrm{std}}/\sigma_{\mathrm{mean}}$
    of the predicted standard deviation across all LOSO held-out trials.
    Activities are sorted by median $F_z$ $\sigma$-CV (descending);
    high values indicate strong within-trial modulation of predicted uncertainty
    (heteroscedasticity), low values indicate near-flat $\sigma$.
    Cyclic, weight-bearing tasks (Stairs, Walking, Footwear) cluster at
    the high-modulation end; quasi-static tasks (Lying, Vibration, Bicycle)
    at the low end. Plotting conventions as in Fig.~\ref{fig:r2_box}.
  }
  \label{fig:sigma_cv}
\end{figure}

\begin{figure}[p]
  \centering
  \includegraphics[width=\linewidth]{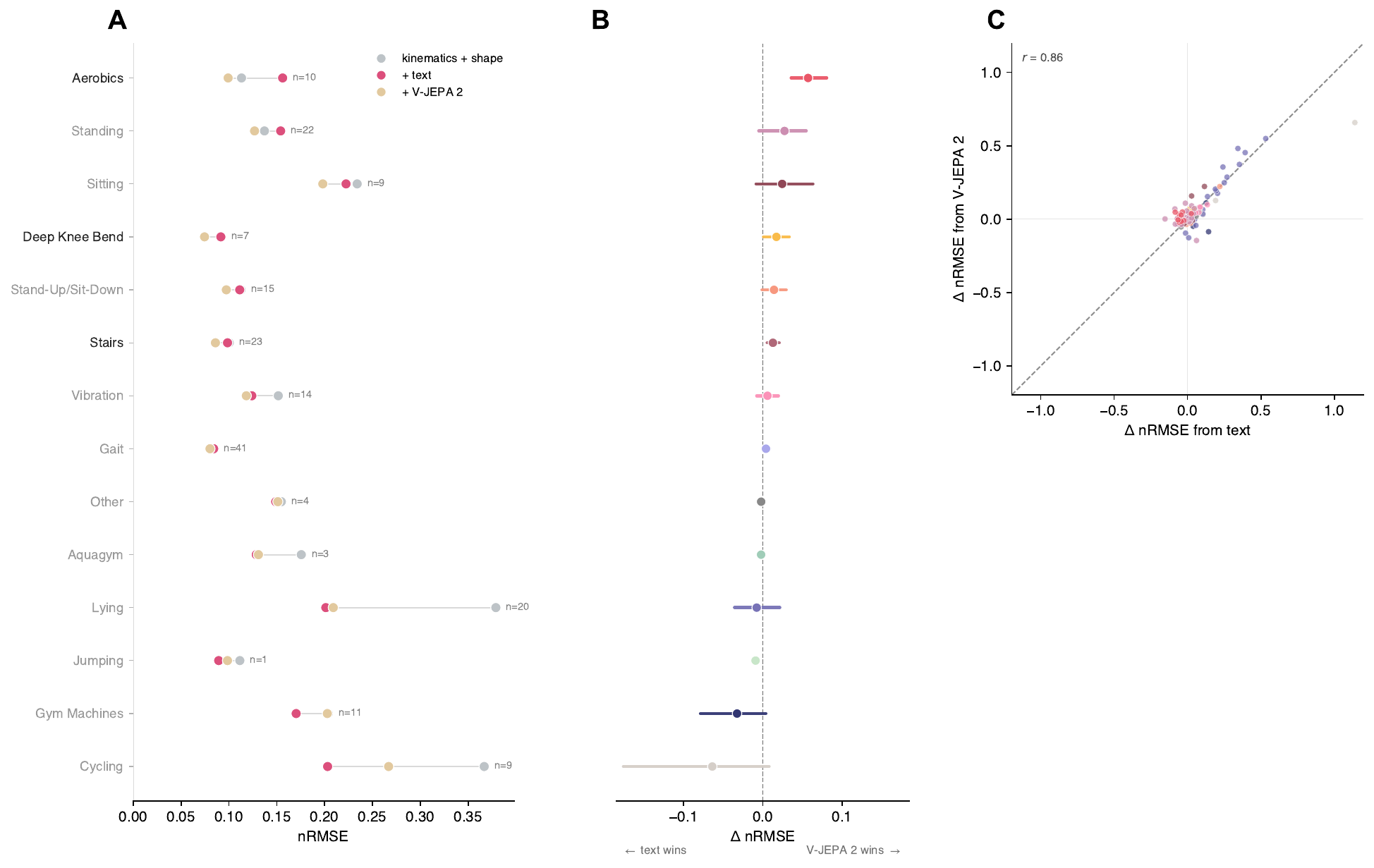}
  \caption{Per-category accuracy of three model variants and the incremental
    benefit of \vjepatwo{} visual features over text embeddings. All panels are computed on the held-out validation split of the 85/15 patient-stratified partition used for the ablations.
    \textbf{(A)}~Mean normalized root mean square error (nRMSE) across the 14
    activity categories present in that split, for the kinematics + shape baseline, the text-augmented
    model, and the \vjepatwo{}-augmented model. Categories are sorted by the mean
    $\text{nRMSE}(\text{+text}) - \text{nRMSE}(\text{+\vjepatwo{}})$ difference
    (ascending), so rows where \vjepatwo{} confers the largest additional benefit
    appear at the top. Activity labels are shown at full contrast for the three
    categories where \vjepatwo{} improves over text (bootstrap 95\%
    CI excludes zero) and grayed otherwise; $n$ denotes the number of trials per
    category.
    \textbf{(B)}~Forest plot of the mean paired difference in nRMSE between the
    text and \vjepatwo{} models (positive values indicate \vjepatwo{} superiority).
    Horizontal bars are 95\% percentile bootstrap confidence intervals
    (2,000 resamples).
    \textbf{(C)}~Per-sample scatter of the improvement in nRMSE conferred by text
    ($x$-axis) versus \vjepatwo{} ($y$-axis) relative to the baseline; the dashed
    line is the identity. Points are colored by activity category, matching the
    colors used in panel~B; per-trial improvements are strongly correlated
    (Pearson $r = 0.86$, $p < 0.001$, $n = 189$ trials).}
  \label{fig:vjepa_breakdown}
\end{figure}

\end{document}